\documentclass[journal]{IEEEtran}

\usepackage{cite}
\usepackage{xcolor}
\usepackage{enumerate}
\usepackage{lineno}
\modulolinenumbers[5]
\usepackage{booktabs}
\usepackage{threeparttable}
\newcommand{\tabincell}[2]{\begin{tabular}{@{}#1@{}}#2\end{tabular}}
\usepackage{booktabs}
\usepackage{caption}
\usepackage[colorlinks=true,citecolor=blue]{hyperref}
\usepackage{subfigure}
\usepackage{multirow}
\usepackage{multicol}
\ifCLASSINFOpdf
  \usepackage[pdftex]{graphicx}
\fi

\usepackage{amsmath}

\usepackage{amsfonts}

\usepackage{algorithm}  
\usepackage{algpseudocode}  
\begin{document}
\title{All Grains, One Scheme (AGOS): 
\\Learning Multi-grain Instance Representation for Aerial Scene Classification}

\author{Qi~Bi,~\IEEEmembership{Student~Member,~IEEE}, Beichen~Zhou, Kun~Qin, Qinghao~Ye,  Gui-Song~Xia,~\IEEEmembership{Senior~Member,~IEEE}% <-this % stops a space

%\thanks{}% <-this % stops a space
\thanks{Qi Bi, Beichen Zhou and Kun Qin are with School of Remote Sensing and Information Engineering, Wuhan University, Wuhan, China. E-mail: {\em qink@whu.edu.cn}.}% <-this % stops a space
%\thanks{Wei Ji is with the Department of Electrical and Computer Engineering, University of Alberta, Edmonton, Canada.}
\thanks{Qinghao Ye is with University of California, San Diego, the United States.}
% <-this % stops a space
\thanks{G.-S. Xia is with the National Engineering Research Center for Multimedia Software, School of Computer Science and Institute of Artificial Intelligence, and also the State Key Lab. LIESMARS, Wuhan University, Wuhan, China. E-mail: {\em guisong.xia@whu.edu.cn}.}
\thanks{Corresponding authors: Qin Kun (qink@whu.edu.cn) and Gui-Song Xia (guisong.xia@whu.edu.cn).}
}
\maketitle

\begin{abstract}
Aerial scene classification remains challenging as: 1) the size of key objects in determining the scene scheme varies greatly; 2) many objects irrelevant to the scene scheme are often flooded in the image. Hence, how to effectively perceive the region of interests (RoIs) from a variety of sizes and build more discriminative representation from such complicated object distribution is vital to understand an aerial scene. 
In this paper, we propose a novel \textit{all grains, one scheme} (AGOS) framework to tackle these challenges. \textit{To the best of our knowledge}, it is the first work to extend the classic multiple instance learning into multi-grain formulation.
Specially, it consists of a multi-grain perception module (MGP), a multi-branch multi-instance representation module (MBMIR) and a self-aligned semantic fusion (SSF) module. Firstly, our MGP preserves the differential dilated convolutional features from the backbone, which magnifies the discriminative information from multi-grains. Then, our MBMIR highlights the key instances in the multi-grain representation under the MIL formulation. Finally, our SSF 
allows our framework to learn the same scene scheme from multi-grain instance representations and fuses them, so that the entire framework is optimized as a whole. 
Notably, our AGOS is flexible and can be easily adapted to existing CNNs in a plug-and-play manner.
Extensive experiments on UCM, AID and NWPU benchmarks demonstrate that our AGOS achieves a comparable performance against the state-of-the-art methods.
\end{abstract}


% \fi
\IEEEpeerreviewmaketitle

\section{Introduction}
\label{sec1}

\IEEEPARstart{A}{erial} scene classification stands at the crossroad of image processing and remote sensing, and has drawn increasing attention in the computer vision community in the past few years \cite{Xia2017DOTA,Bi2019Multiple,Ding2019Learning,Mou2019Relation,Bi2021LSENet}. Moreover, aerial scene classification is a fundamental task towards the understanding of aerial images, as it plays a significant role on many aerial image applications such as land use classification \cite{Tong2019Land,Bi2019A,Hong2020XModal} and urban planning \cite{Zhang2020Land}. 

\subsection{Problem Statement}

Despite the great performance gain led by deep learning for image recognition \cite{Gao2017Densely,ResNet,VGGNet,GoogLeNet,Ji2021SOD,Ji2021WSOD}, aerial scene classification remains challenging due to some unique characteristics:
\begin{figure}[htb!]
    \vspace{-2mm}
    \centering %插入的图片居中表示
    \subfigure[average object size]{
    \includegraphics[width=1.6in]{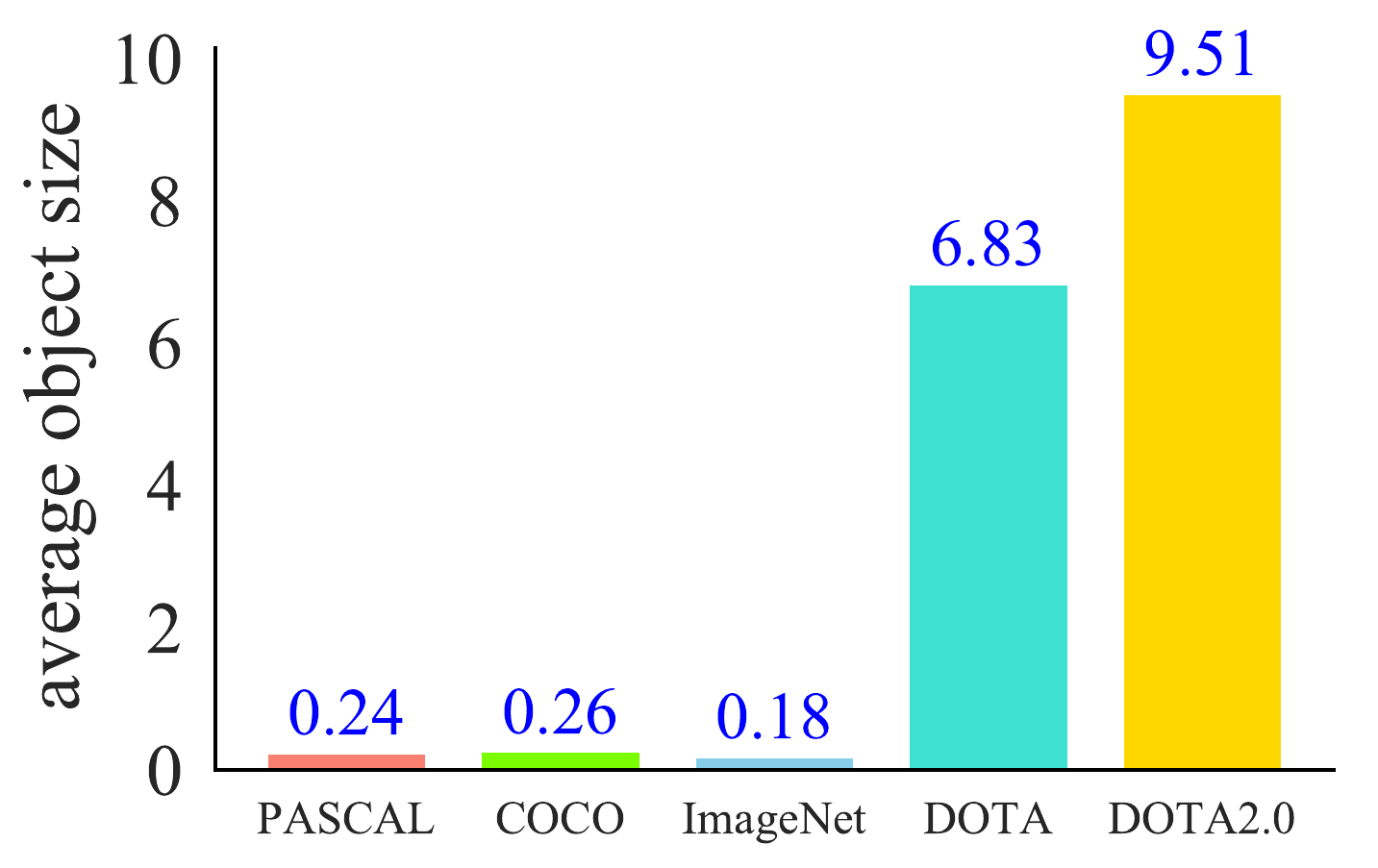}
    }
    \subfigure[average object quantity]{
    \includegraphics[width=1.6in]{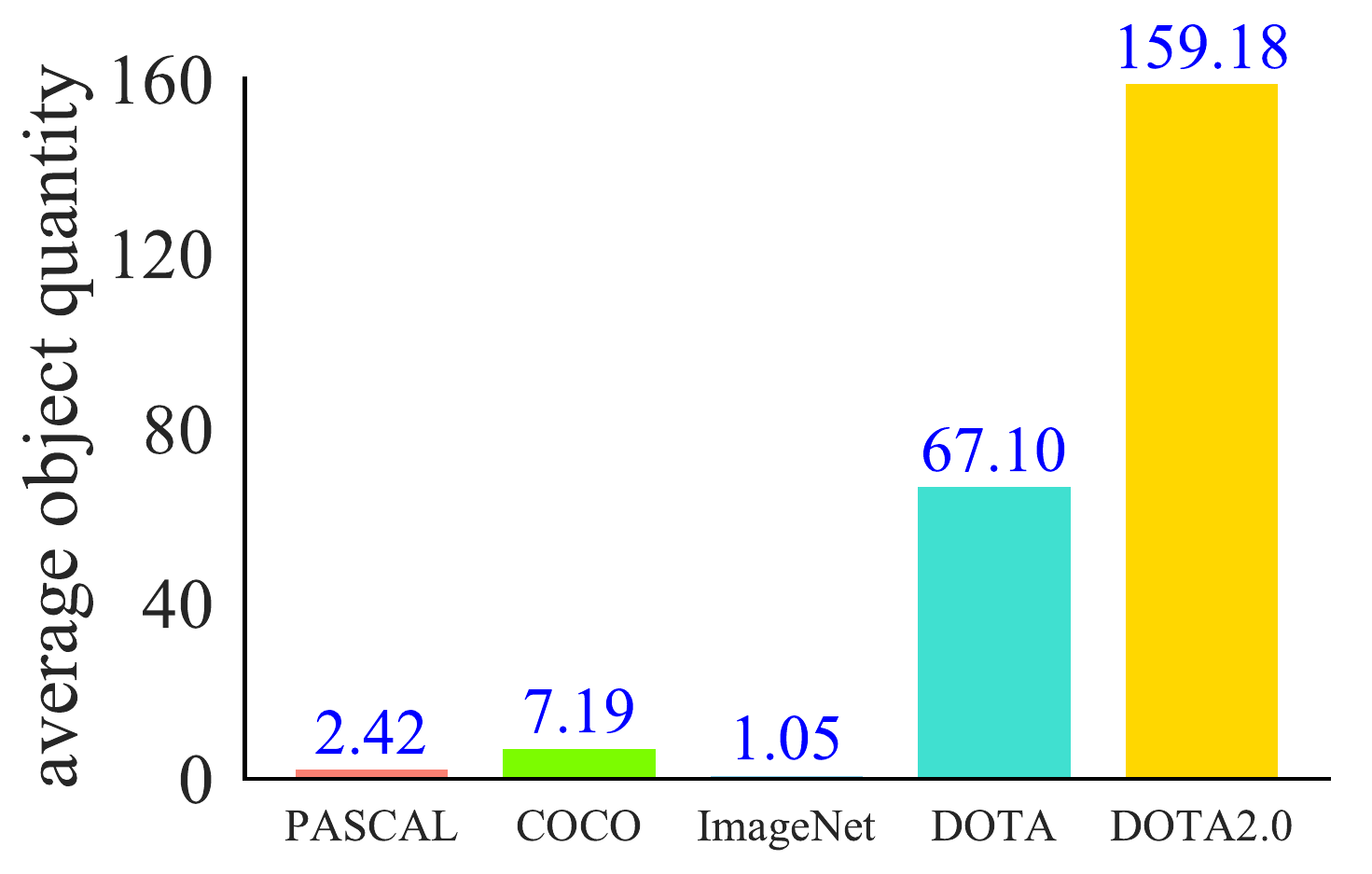}
    }
    
    \subfigure[Example on object size]{
    \includegraphics[width=1.6in]{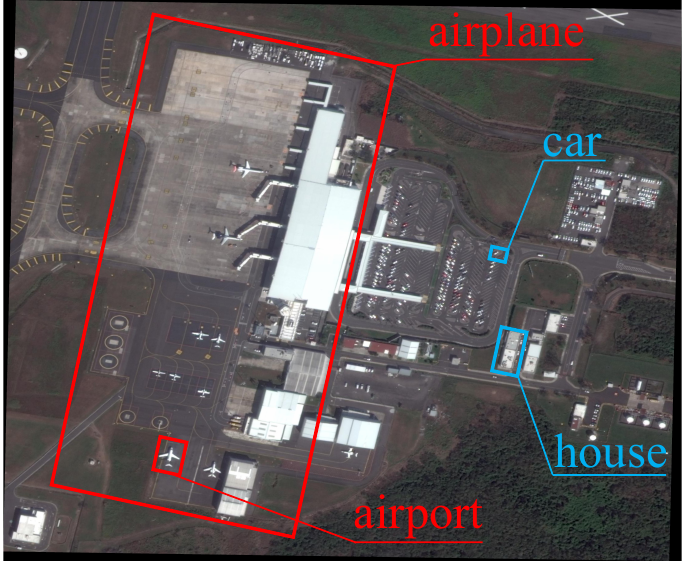}
    }
    \subfigure[Example on object quantity]{
    \includegraphics[width=1.6in,height=1.31in]{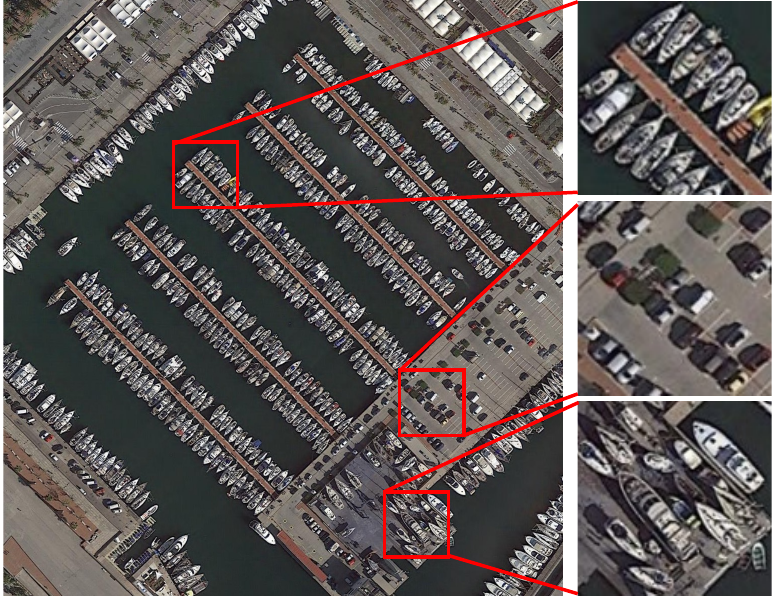}
    }
    %插入的图，包括JPG,PNG,PDF,EPS等，放在源文件目录下
	\caption{Different statistics between aerial image samples (from \textit{DOTA} and \textit{DOTA2.0}) and ground image samples (from \textit{PASCAL}, \textit{COCO} and \textit{ImageNet}) on (a) average object sizes and (b) average object quantity. All the original statistics are quoted from \cite{DOTA2.0}. It can be clearly seen that objects from aerial images are much more varied in sizes and each aerial image usually has much more objects. (c) \& (d): Example on the dramatically varied object size and huge object amount in aerial images.}  %图片的名称
	%\vspace{-1.0em}
	\label{fig1}   %标签，用作引用
\end{figure}

\textit {1) More varied object sizes in aerial images.}
As both the spatial resolution and viewpoint of the sensor vary greatly in aerial imaging \cite{Xia2017AID,He2018Remote,Xia2017DOTA}, the object size from bird view is usually more varied compared with the ground images. Specifically, the objects in ground images are usually middle-sized. In contrast, there are much more small-sized objects in aerial images but some of the objects such as airport and roundabout are extremely large-sized. As a result, the average object size from aerial images is much higher than the ground images (shown in Fig.~\ref{fig1}~(a) \& (c)).

Thus, it is difficult for existing convolutional neural networks (CNNs) with a fixed receptive field to fully perceive the scene scheme of an aerial image due to the more varied sizes of key objects \cite{Hu2015Transferring,Han2017Pre,Li2017Integrating,Xia2017DOTA,Bi2021LSENet}, which pulls down the understanding capability of a model for aerial scenes. 

\textit {2) More crowded object distribution in aerial images.} Due to the bird view from imaging platforms such as unmanned aerial vehicles and satellites, the aerial images are usually large-scale and thus contain much more objects than ground images \cite{Bi2019Multiple,Xia2017DOTA,FARelation2020} (see Fig.~\ref{fig1}~(b) \& (d) for an example).

\begin{figure}[t!]
    \centering %插入的图片居中表示
	\includegraphics[width=3.4in]{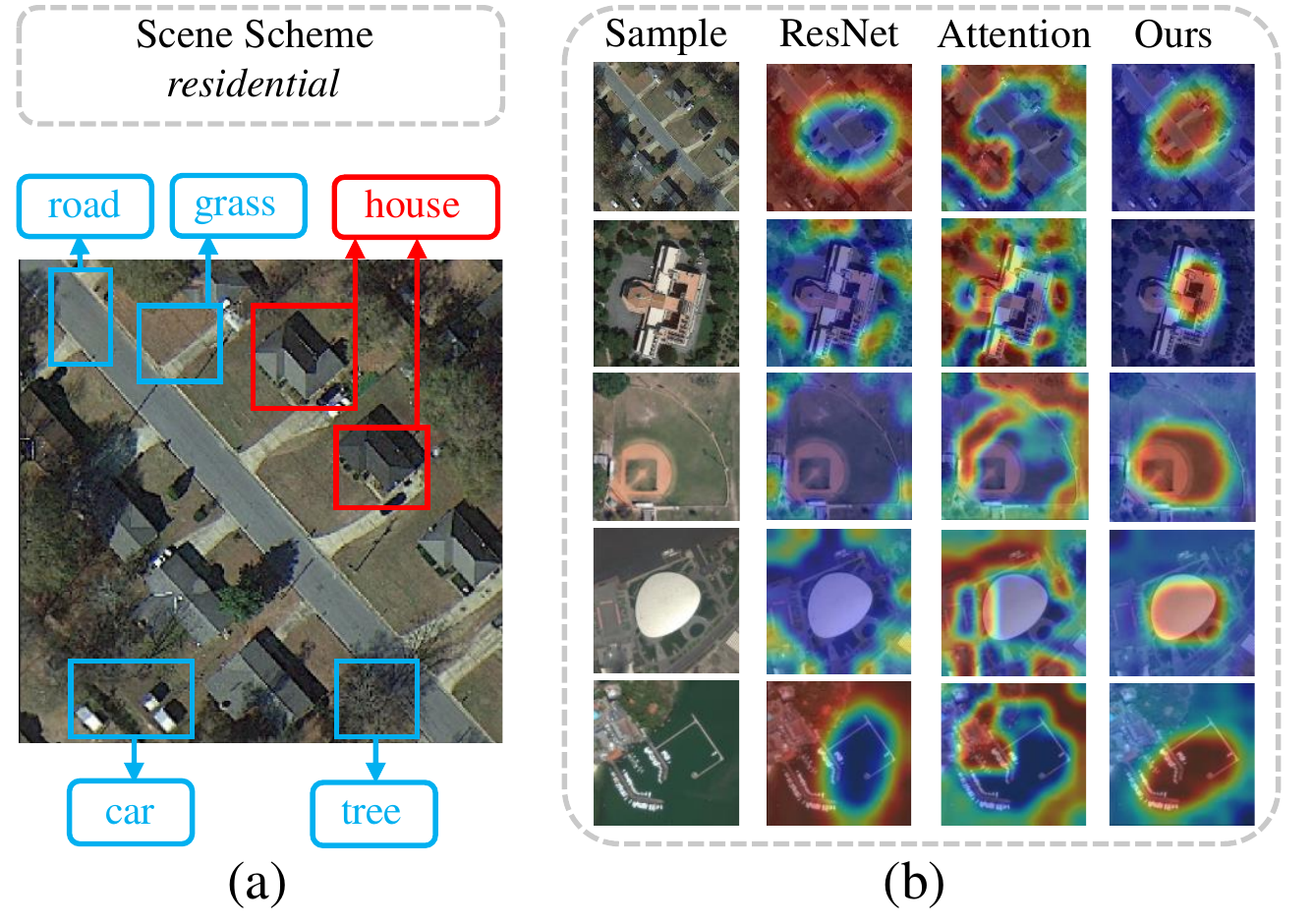} %插入的图，包括JPG,PNG,PDF,EPS等，放在源文件目录下
	\caption{An intuitive illustration on how aerial scenes contain more objects irrelevant to the scene scheme (a) and existing CNNs can fail to activate the RoIs in aerial scenes (b). In (a), local regions relevant and irrelevant to the scene scheme are labeled in red and blue respectively.}  %图片的名称
	\vspace{-1.0em}
	\label{fig2}   %标签，用作引用
\end{figure}

Unfortunately, existing CNNs are capable of preserving the global semantics \cite{ResNet,VGGNet,GoogLeNet} but are unqualified to highlight the key local regions \cite{Ilse2018Attention,Wieland2019Approximating}, \textit{i.e.}, region of interests (RoIs), of a scene with complicated object distributions. Therefore, CNNs are likely to be affected by the local semantic information irrelevant to the scene label and fail to predict the correct scene scheme \cite{Bi2019Multiple,WangQi2018RA,Zhu2018Adaptive,Penatti2015Do,Gong2018When} (see Fig.~\ref{fig2}~for an intuitive illustration).
%\vspace{-2.5em}

\subsection{Motivation \& Objectives}

We are motivated to tackle the above challenges in aerial scene classification, hoping to build a more discriminative aerial scene representation. Specific objectives include:

\textit {1) Highlighting the key local regions in aerial scenes.} Great effort is needed to highlight the key local regions of an aerial scene for existing deep learning models, so as to correctly perceive the scene scheme rather than activate the background or other local regions in an aerial scene. 

Therefore, the formulation of classic multiple instance learning (MIL) \cite{Dietterich1997Solving,Maron1998Multiple} is adapted in our work to describe the relation between the aerial scene (bag) and the local image patches (instances). This formulation helps highlight the feature responses of key local regions, and thus enhances the understanding capability for the aerial scene.

\textit {2) Aligning the same scene scheme for multi-grain representation.} Allowing for the varied object sizes in an aerial scene, it is natural to use existing multi-scale convolutional features \cite{Han2017Pre,Hu2015Transferring,Li2017Integrating,He2018Remote} for more discriminative aerial scene representation. However, given the aforementioned complicated object distribution in the aerial scene, whether the representation 
of each scale learnt from existing multi-scale solutions can focus on the scene scheme remains to be an open question but is crucial to depict the aerial scenes.

Hence, different from existing multi-scale solutions \cite{Pang2019towards}, we extend the classic MIL formulation to a multi-grain manner under the existing deep learning pipeline, in which a set of instance representations are built from multi-grain convolutional features. More importantly, in the semantic fusion stage, we develop a simple yet effective strategy to align the instance representation from each grain to the same scene scheme. 

%\vspace{-1.0em}
\subsection{Contribution}

To realize the above objectives, our contribution in this paper can be summarized as follows. 
\begin{enumerate}[(1)]
\item We propose an \textit{all grains, one scheme} (AGOS) framework for aerial scene classification. \textit{To the best of our knowledge}, we are the first to formulate the classic MIL into deep multi-grain form. Notably, our framework can be adapted into the existing CNNs in a plug-and-play manner.
\item We propose a bag scheme self-alignment strategy, which allows the instance representation from each grain to highlight the key instances corresponding to the bag scheme without additional supervision. Technically, it is realized by our self-aligned semantic fusion (SSF) module and semantic-aligning loss function. 
\item We propose a multi-grain perception (MGP) module for multi-grain convolutional feature extraction. Technically, the absolute difference from each two adjacent grains generates more discriminative aerial scene representation. 
\item Extensive experiments not only validate the state-of-the-art performance of our AGOS on three aerial scene classification benchmarks, but also demonstrate the generalization capability of our AGOS on a variety of CNN backbones and two other classification domains.
\end{enumerate}

This paper is an extension of our conference paper accepted by the ICASSP 2021 \cite{ICASSP2021}. Compared with \cite{ICASSP2021}, the specific improvement of this paper includes: 1) The newly-designed bag scheme self-alignment strategy, realized by our SSF module and the corresponding loss function, is capable to align the bag scheme to the instance representation from each grain; 2) We design a multi-grain perception module, which additionally learns the \textit{base instance representation}, to align the bag scheme and to highlight the key local regions in aerial scenes; 3) Empirically, our AGOS demonstrates superior performance of against our initial version \cite{ICASSP2021}. Also, more experiments, discussion and visualization are provided to analyze the insight of our AGOS. 

The remainder of this paper is organized as follows. In Section~\ref{sec2}, related work is provided. In Section~\ref{sec3}, the proposed method is demonstrated. In Section~\ref{sec4}, we report and discuss the experiments on three aerial image scene classification benchmarks. Finally in Section~\ref{sec5}, the conclusion is drawn.

\section{Related work}
\label{sec2}

\subsection{Aerial scene classification}

Aerial scene classification remains a heated research topic for both the computer vision and the remote sensing community. In terms of the utilized features, these solutions are usually divided into the low-level (\textit{e.g.}, color histogram \cite{Jae2012Color}, wavelet transformation \cite{Bin2008Indexing}, local binary pattern \cite{Amit2014LBP,Ojala2002Multiresolution} and \textit{etc.}), middle-level (\textit{e.g.}, bag of visual words \cite{Zhong2015Scene}, potential latent semantic analysis \cite{Hofmann2001Unsupervised,Blei2003Latent}, latent dirichlet allocation \cite{Zhao2015Dirichlet} and \textit{etc.}) and high-level feature based methods.  

High-level feature methods, also known as deep learning methods, have become the dominant paradigm for aerial scene classification in the past decade. Major reasons accounting for its popularity include their stronger feature representation capability and end-to-end learning manner \cite{Zhu2017DeepR,Cheng2020Land}.

Among these deep learning based methods, CNNs are the most commonly-utilized \cite{Bi2019Multiple,Hu2015Transferring,Han2017Pre,Nogueira2017Towards,Li2017Integrating,He2018Remote} as the convolutional filters are effective to extract multi-level features from the image. In the past two years, CNN based methods (e.g., DSENet \cite{Wang2021Enhanced}, MS2AP \cite{Bi2021MS}, MSDFF \cite{Xue2020Multi}, CADNet \cite{Tong2020Channel}, LSENet \cite{Bi2021LSENet}, GBNet \cite{Sun2020GBN}, MBLANet \cite{Chen2021MBLANet}, MG-CAP \cite{Wang2020MGCAP}, Contourlet CNN \cite{Liu2021CCNN}) still remain heated for aerial scene classification. On the other hand, recurrent neural network (RNN) based \cite{WangQi2018RA}, auto-encoder based \cite{Cheng2015Effective,Zhang2014Saliency} and generative adversarial network (GAN) based \cite{Lin2017MARTA,Yu2019AGAN} approaches have also been reported effective for aerial scene classification.

Meanwhile, although recently vision transformer (ViT)  \cite{2021CViT,2021TRS,2021ViTRS} have also been reported to achieve high classification performance for remote sensing scenes, as they focus more on the global semantic information with the self-attention mechanism while our motivation focus more on the local semantic representation and activation of region of interests (RoIs). Also, the combination of multiple instance learning and deep learning is currently based on the CNN pipelines \cite{Ilse2018Attention,Bi2019Multiple,Shi2020loss,Shi2021loss,Wang2016Revisiting}. Hence, the discussion and comparison of ViT based methods are beyond the scope of this work. 

To sum up, as the global semantic representation of CNNs is still not capable enough to depict the complexity of aerial scenes due to the complicated object distribution \cite{Bi2019Multiple,WangQi2018RA}, how to properly highlight the region of interests (RoIs) from the complicated background of aerial images to enhance the scene representation capability still remains rarely explored.

\subsection{Multi-scale feature representation}

Multi-scale convolutional feature representation has been long investigated in the computer vision community \cite{He2015Spatial,Lin2017FPN}. As the object sizes are usually more varied in aerial scenes, multi-scale convolutional feature representation has also been widely utilized in the remote sensing community for a better understanding of aerial images. 

Till now, multi-scale feature representation for aerial images can be classified into two categories, that is, using multi-level CNN features in a non-trainable manner and directly extracting multi-scale CNN features in the deep learning pipeline.

For the first category, the basic idea is to derive multi-layer convolutional features from a pre-trained CNN model, and then feed these features into a non-trainable encoder such as BoW or LDA. Typical works include \cite{Hu2015Transferring,Li2017Integrating,Nogueira2017Towards}. Although the motivation of such approaches is to learn more discriminative scene representation in the latent space, they are not end-to-end and the performance gain is usually marginal. 

For the second category, the basic idea is to design spatial pyramid pooling \cite{Han2017Pre,Bi2021MS} or image pyramid \cite{He2018Remote} to extend the convolutional features into multi-scale representation. Generally, such multi-scale solutions can be further divided into four categories \cite{Pang2019towards}, namely, encoder-decoder pyramid, spatial pyramid pooling, image pyramid and parallel pyramid. 

Although nowadays multi-scale representation methods become mature, whether the representation from each scale can effectively activate the RoIs in the scene has not been explored. 
%To solve the bottleneck of these solutions, we intend to perceive the multi-grain responses from the same convolutional feature map while highlighting the key local regions. 

\begin{figure*}[!t]
    \centering %插入的图片居中表示
	\includegraphics[width=6.8in]{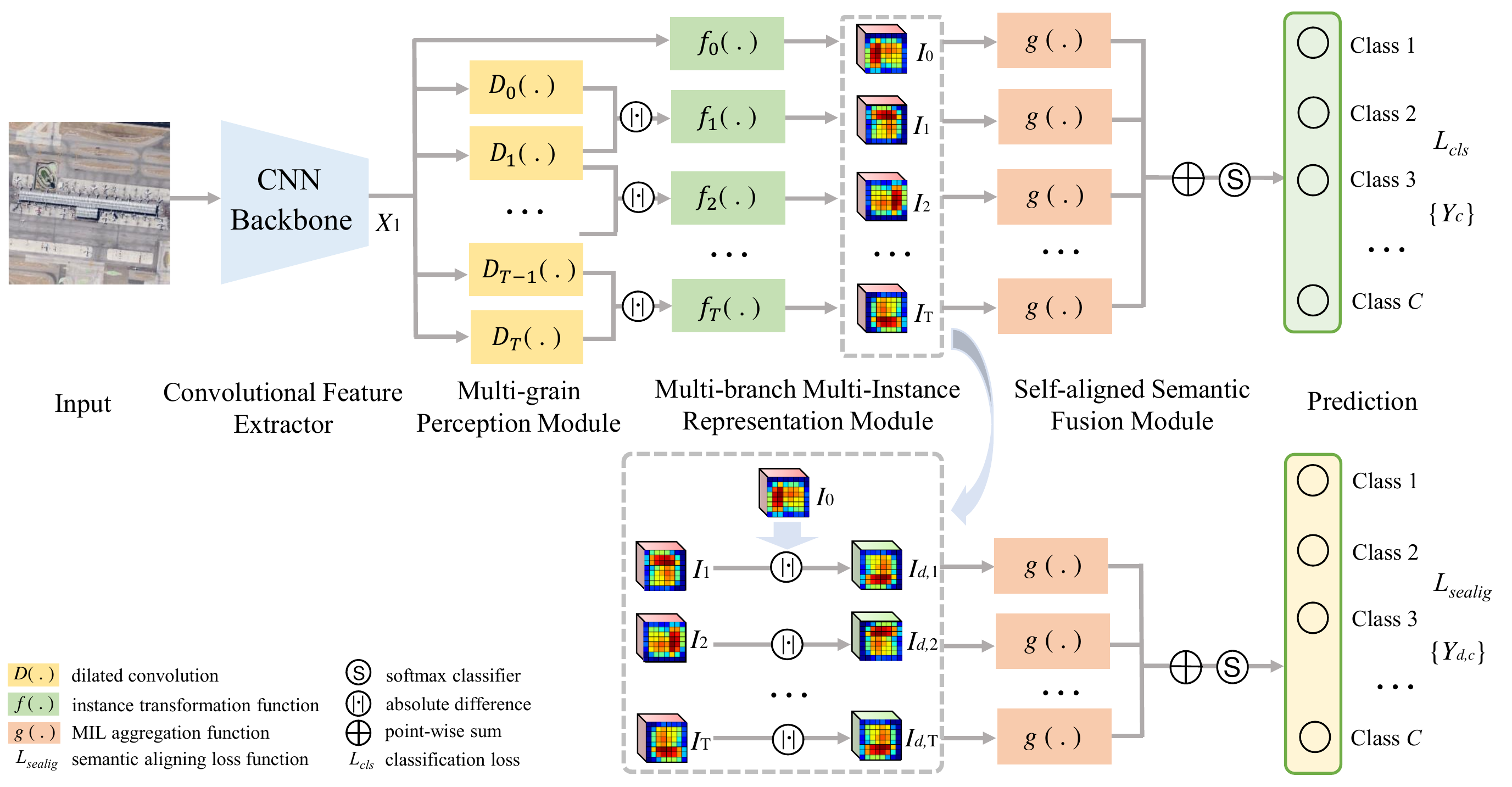} %插入的图，包括JPG,PNG,PDF,EPS等，放在源文件目录下
	\caption{Our proposed \textit{all grains, one scheme} (AGOS) framework.}  %图片的名称
	\label{fig3}   %标签，用作引用
\end{figure*}

\subsection{Multiple instance learning}

Multiple instance learning (MIL) was initially designed for drug prediction \cite{Dietterich1997Solving} and then became an important machine learning tool \cite{Maron1998Multiple}. In MIL, an object is regarded as a bag, and a bag consists of a set of instances \cite{Zhang2004Improve}. Generally speaking, there is no specific instance label and each instance can only be judged as either belonging or not belonging to the bag category. This formulation makes MIL especially qualified to learn from the weakly-annotated data \cite{Pinheiro2015From,Tang2017Learning,Wang2016Revisiting}.

The effectiveness of MIL has also been validated on a series of computer vision tasks such as image recognition \cite{Wang2013MILDic}, saliency detection \cite{Wang2013Saliency,Zhang2016Co}, spectral-spatial fusion \cite{Liu2018PAN} and object localization/detection \cite{Wang2015Large,Zhu2017Deep,Mou2018Vehicle,Wang2015MISVM,Tang2017MILde}. 

On the other hand, the classic MIL theory has also been enriched. Specifically, Sivan \textit{et al.} \cite{JMLR12} relaxed the Boolean OR assumption in MIL formulation, so that the relation between bag and instances becomes more general. More recently, Alessandro \textit{et al.} \cite{JMLR20} investigated a three-level multiple instance learning. The three hierarchical levels are in a vertical manner, and they are top-bag, sub-bag, instance, where the sub-bag is an embedding between the top-bag and instances. Note that our deep MIL under multi-grain form is quite distinctive from \cite{JMLR20} as our formulation still has two hierarchical levels, \textit{i.e.}, bag and instances, and the instance representation is generated from multi-grain features.

In the past few years, deep MIL draws some attention, in which MIL has the trend to be combined with deep learning in a trainable manner. To be specific, Wang \textit{et al.} utilized either max pooling or mean pooling to aggregate the instance representation in the neural networks \cite{Wang2016Revisiting}. Later, Ilse \textit{et al.} \cite{Ilse2018Attention} used a gated attention module to generate the weights, which are utilized to aggregate the instance scores. Bi \textit{et al.} \cite{Bi2019Multiple} utilized both spatial attention module and channel-spatial attention module to derive the weights and directly aggregate the instance scores into bag-level probability distribution. More recently, Shi \textit{et al.} \cite{Shi2020loss,Shi2021loss} embedded the attention weights into the loss function so as to guide the learning process for deep MIL.

\begin{table}[!t]  
    \centering
    \caption{A brief summary on the attributes of our AGOS and recent deep multiple instance learning algorithms. For space attribute, $I$: instance-space paradigm, $E$: Embedding-space paradigm; Aggregate attribute refers to the MIL aggregation function; Grain, $S$: single-grain, $M$: multi-grain. }
    \begin{tabular}{ccccc} 
    \hline
    Method & Space & Aggregation & Hierarchy & Grain \\ 
    \hline
    \cite{Wang2016Revisiting} & $I,E$ & max/mean & bag-instance & $S$ \\
    \cite{Ilse2018Attention} & $E$ & attention & bag-instance & $S$ \\
    \cite{Bi2019Multiple} &	$I$ & attention & bag-instance & $S$ \\
    \cite{Shi2020loss,Shi2021loss} & $I$ & attention & bag-instance & $S$ \\
    \cite{JMLR20} & $E$ & max & top bag-bag-instance & $S$ \\
    \hline
    AGOS (Ours) & $I$ & mean & bag-instance & $M$ \\
    \hline
    \end{tabular} 
    \label{tabMIL}
\end{table}

\section{The proposed method}
\label{sec3}
\subsection{Preliminary}
\label{sec3.1}
\textit{1) Classic \& deep MIL formulation:} For our aerial scene classification task, according to the classic MIL formulation \cite{Maron1998Multiple,Dietterich1997Solving}, a scene $X$ is regarded as a bag, and the bag label $Y$ is the same as the scene category of this scene. As each bag $X$ consists of a set of instances $\{x_{1}, x_{2}, \cdots, x_{l}\}$, each image patch of the scene is regarded as an instance.

All the instances indeed have labels $y_1, y_2, \cdots, y_l$, but all these instance labels are weakly-annotated, \textit{i.e.}, we only know each instance either belongs to (denoted as 1) or does not belong to (denoted as 0) the bag category. Then, whether or not a bag belongs to a specific category $c$ is determined via
\begin{equation}\label{eq1}
Y=
\begin{cases}
0 & \text{if} \sum_{t=1}^l y_t = 0 \\
1 & \text{else}
\end{cases} .
\end{equation}

In deep MIL, as the feature response from the gradient propagation is continuous, the bag probability prediction $Y$ is assumed to be continuous in $[0,1]$ \cite{Ilse2018Attention,Bi2019Multiple}. It is determined to be a specific category $c$ via
\begin{equation}\label{eq2}
Y=
\begin{cases}
1& \text{if $p_{c}= \max \{p_{1}, \cdots, p_{C}\}$}\\
0& \text{else}
\end{cases},
\end{equation}
where $p_{1}, p_{2}, \cdots, p_{c}, \cdots, p_{C}$ denotes the bag probability prediction of all the total $C$ bag categories. 

\textit{2) MIL decomposition:} In both classic MIL and deep MIL, the transition between instances $\{x_{s}\}$ (where $s=1,2,\cdots,l$) to the bag label $Y$ can be presented as 
\begin{equation} \label{eq3}
Y=h\Big( g\big( f(\{x_{s}\}) \big) \Big),
\end{equation}
where $f$ denotes a transformation which converts the instance set into an instance representation, $g$ denotes the MIL aggregation function, and $h$ denotes a transformation to get the bag probability distribution. 

\textit{3) Instance space paradigm:} The combination of MIL and deep learning is usually conducted in either instance space \cite{Wang2016Revisiting,Bi2019Multiple,Shi2021loss} or embedding space \cite{Ilse2018Attention}. Embedding space based solutions offer a latent space between the instance representation and bag representation, but this latent space in the embedding space can sometimes be less precise in depicting the relation between instance and bag representation \cite{Ilse2018Attention,Bi2019Multiple}. In contrast, instance space paradigm has the advantage to generate the bag probability distribution directly from the instance representation \cite{Wang2016Revisiting,Bi2019Multiple}. Thus, the $h$ transformation in Eq.~\ref{eq3}~becomes an identity mapping, and it is rewritten as 
\begin{equation} \label{eq4}
Y=g(f(\{x_{s}\})).
\end{equation}
\vspace{-1.5em}

\textit{4) Problem formulation:} As we extend MIL into multi-grain form, the transformation function $f$ in Eq.~\ref{eq4}~is extended to a set of transformations $\{f_{t}\}$ (where $t=1,2,\cdots,T$). Then, $Y$ is generated from all these grains and thus Eq.~\ref{eq4} can be presented as 
\begin{equation} \label{eq5}
Y=g(f_{1}(\{x_{s}\}),f_{2}(\{x_{s}\}),\cdots,f_{T}(\{x_{s}\})).
\end{equation}
\vspace{-1.5em}

Hence, how to design a proper and effective transformation set $\{f_{t}\}$ and the corresponding MIL aggregation function $g$ under the existing deep learning pipeline is our major task.

\textit{5) Objective:} Our objective is to classify the input scene $X$ in the deep learning pipeline under the formulation of multi-grain multi-instance learning. To summarize, the overall objective function can be presented as
\begin{equation} \label{eq6}
\mathop{\arg\min}\limits_{W,b}{ \mathcal{L}(Y,g(f_{1}(\{x_{s}\}),\cdots,f_{T}(\{x_{s}\});W,b))\\+\Psi(W)},
\end{equation}
where $W$ and $b$ is the weight and bias matrix to train the entire framework, $\mathcal{L}$ is the loss function and $\Psi$ is the regularization term. 

Moreover, how the instance representation of each grain $f_{t}(\{x_{s}\})$ is aligned to the same bag scheme is also taken into account in the stage of instance aggregation $g$ and optimization $\mathcal{L}$, which can be generally presented as
\begin{equation} \label{eq7}
\begin{aligned}
s.t. \quad g(f_{1}(\{x_{s}\}))=g(f_{2}(\{x_{s}\}))=\cdots \\
=g(f_{t}(\{x_{s}\}))=\cdots=g(f_{T}(\{x_{s}\}))=Y_{c},
\end{aligned}
\end{equation}
where $Y_{c}$ denotes the category that the bag belongs to. 
%\vspace{-1.5em}
\subsection{Network overview}
\label{sec3.2}
As is shown in Fig.~\ref{fig3}, our proposed \textit{all grains, one scheme} (AGOS) framework consists of three components after the CNN backbone. To be specific, the multi-grain perception module (in Sec.~\ref{sec3.3}) implements our proposed differential dilated convolution on the convolutional features so as to get a discriminative multi-grain representation. Then, the multi-grain feature presentation is fed into our multi-branch multi-instance representation module (in Sec.~\ref{sec3.4}), which converts the above features into instance representation, and then directly generates the bag-level probability distribution. 
As aligning the instance representation from each grain to the same bag scheme is another important objective, we  propose a bag scheme self-alignment strategy, which is technically fulfilled by our self-aligned semantic module (in Sec.~\ref{sec3.5}) and the corresponding loss function (in Sec.~\ref{sec3.6}). In this way, the entire framework is trained in an end-to-end manner.
%\vspace{-1.5em}

\subsection{Multi-grain Perception Module}
\label{sec3.3}
\begin{figure}[!t]
    \centering %插入的图片居中表示
	\includegraphics[width=3.4in]{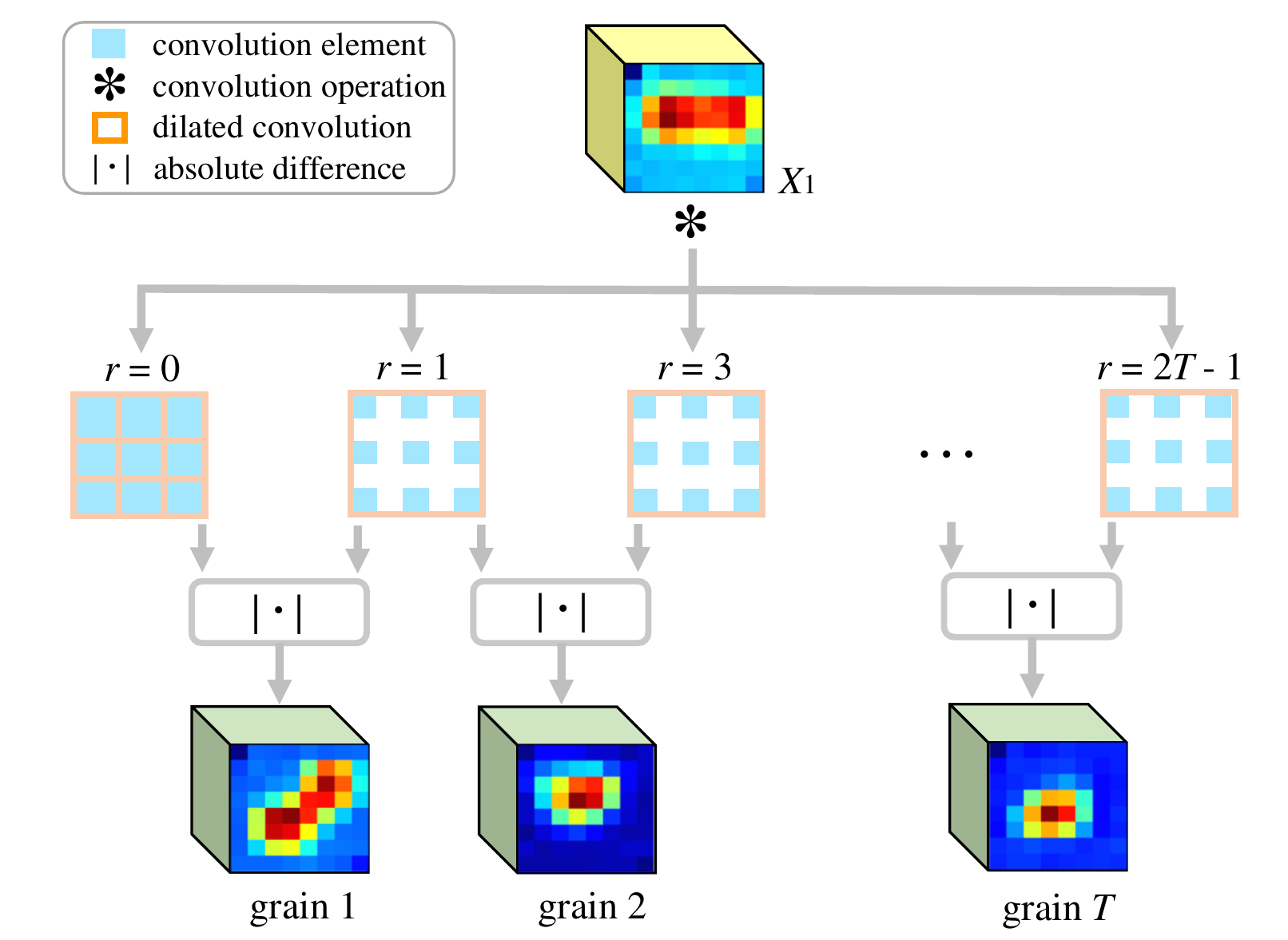} %插入的图，包括JPG,PNG,PDF,EPS等，放在源文件目录下
	\caption{Demonstration on our multi-grain perception (MGP) module. The inputted convolutional features are processed by a set of dilated convolution with different dilated rate. Then, absolute difference of each adjacent representation pairs is calculated to get the multi-grain representation output.}  %图片的名称
	\label{fig4}   %标签，用作引用
\end{figure}

\textit{1) Motivation:} Our multi-grain perception (MGP) module intends to convert the convolutional feature from the backbone to multi-grain representations. Different from existing multi-scale strategies \cite{Hu2015Transferring,Han2017Pre,Li2017Integrating,He2018Remote}, our module builds same-sized feature maps by perceiving multi-grain representations from the same convolutional feature. Then, the absolute difference of the representations from each two adjacent grains is calculated to highlight the differences from a variety of grains for more discriminative representation (shown in Fig.~\ref{fig4}).

\textit{2) Dilated convolution:} Dilated convolution is capable of perceiving the feature responses from different receptive field while keeping the same image size \cite{Chen2018DeepLab}. Thus, it has been widely utilized in many visual tasks in the past few years. 

Generally, dilation rate $r$ is the parameter to control the window size of a dilated convolution filter. For a $3\times3$ convolution filter, a dilation rate $r$ means that $r-1$ zero-valued elements will be padded into two adjacent elements of the convolution filter. For example, for a $3\times3$ convolution filter, a dilation rate will expand the original convolutional filter to the size of $(2r+1)\times(2r+1)$. Specifically, when $r=0$, there is no zero padding and the dilated convolutional filter degrades into the traditional convolution filter. 

\textit{3) Multi-grain dilated convolution:} Let the convolutional feature from the backbone denote as $X_{1}$. Assume there are $T$ grains in our MGP, then $T$ dilated convolution filters are implemented on the input $X_{1}$, which we denote as $D_{1}, D_{2}, \cdots, D_{T}$ respectively. Apparently, the set of multi-grain dilated convolution feature representation $X_{1}^{'}$ from the input $X_{1}$ can be presented as
\begin{equation} \label{eq8}
X_{1}^{'}=\{X_{1}^{1},X_{1}^{2},\cdots,X_{1}^{T}\},
\end{equation}
where we have
\begin{equation} \label{eq9}
X_{1}^{t}=D_{t}(X_{1}),
\end{equation}
and $t=1, 2, \cdots, T$.

The determination of the dilation rate $r$ for the multi-grain dilated convolution set $\{D_{t}\}$ follows the existing rules \cite{Chen2018DeepLab} that $r$ is set as an odd value, \textit{i.e.,} $r=1, 3, 5,\cdots$. Hence, for $D_{t}$, the dilation rate $r$ is $2t-1$.

\textit{4) Differential dilated convolution:} To reduce the feature redundancy from different grains while stressing the discriminative features that each grain contains, absolute difference of each two adjacent representations in $X_{1}^{'}$ is calculated via
\begin{equation} \label{eq10}
\begin{aligned}
%& X_{d,1}=\|D_{1}(X_{1})-D_{0}(X_{1})\|_{1}, \\
%& X_{d,2}=\|D_{2}(X_{1})-D_{1}(X_{1})\|_{1}, \\
%& \quad \quad \cdots \\
%& 
X_{d,t}=\|D_{t}(X_{1})-D_{t-1}(X_{1})\|, \\
%& \quad \quad \cdots \\
%& X_{d,T}=\|D_{T}(X_{1})-D_{T-1}(X_{1})\|_{1}, \\
%\setlength{\belowdisplayskip}{1pt}
\end{aligned}
\end{equation}
where $\| \cdot \|$ denotes the absolute difference, and $X_{d,t}$ ($t=1, 2, \cdots, T$) denotes the calculated differential dilated convolutional feature representation. It is worth noting that when $t=1$, $D_{0}(X_{1})$ means the dilated convolution degrades to the conventional convolution.

Finally, the output of this MGP module is a set of convolutional feature representation $X_{1}^{''}$, presented as
\begin{equation} \label{eq11}
X_{1}^{''}=\{X_{d,0},X_{d,1},X_{d,2},\cdots,X_{d,T}\},
\end{equation}
where $X_{d,0}$ denotes the base representation in our bag scheme self-alignment strategy, the function of which will be discussed in detail in the next two subsections.

Generally, $X_{d,0}$ is a direct refinement of the input $X_{1}$ in the hope of highlighting the key local regions. The realization of this objective is straight forward, as the $1\times1$ convolutional layer has recently been reported to be effective in refining the feature map and highlight the key local regions \cite{Gao2017Densely,Bi2019Multiple}. This process is presented as
\begin{equation} \label{eq12}
X_{d,0}^{W\times H \times C_{1}}=W_{d,0}^{W\times H \times C_{1}}X_{1}^{W\times H \times C_{1}}+b_{d,0}^{W\times H \times C_{1}},
\end{equation}
where $W_{d,0}$ and $b_{d,0}$ denotes the weight and bias matrix of this $1\times1$ convolutional layer, $W$ and $H$ denotes the width and height of the feature representation $X_{1}$. Moreover, as the channel number $C_{1}$ of $X_{d,0}$ keeps the same with $X_{1}$, so the number of convolutional filters in this convolutional layer also equals to the above channel number $C_{1}$. 

\textit{5) Summary:} As shown in Fig.~\ref{fig4} and depicted from Eq.~\ref{eq8} to \ref{eq12}, in our MGP, the inputted convolutional features are processed by a series of dilated convolution with different dilated rate. Then, the absolute difference of each representation pair from the adjacent two grains (\textit{i.e., $r=1$ and $r=3$, $r=3$ and $r=5$}) is calculated as output, so as to generate the multi-grain differential convolutional features for more discriminative representation.

\subsection{Multi-branch Multi-instance Representation Module}
\label{sec3.4}

\textit{1) Motivation:} The convolutional feature representations $X_{1}^{''}$ from different grains contain different discriminative information in depicting the scene scheme. Hence, for the representation $X_{d,t}$ from each grain ($t=1, 2,\cdots, T$), a deep MIL module is utilized to highlight the key local regions. Specifically, each module converts the convolutional representation into an instance representation, and then utilizes an aggregation function to get the bag probability distribution. All these parallel modules are organized as a whole for our multi-branch multi-instance representation (MBMIR) module.

\textit{2) Instance representation transformation:} Each convolutional representation $X_{d,t}$ (where $t=0, 1, \cdots, T$) in the set $X_{1}^{''}$ needs to be converted into an instance representation by a transformation at first, which is exactly the $f$ function in Eq.~\ref{eq3}~and~\ref{eq4}. Specifically, for $X_{d,t}$, this transformation can be presented as
\begin{equation} \label{eq13}
I_{t}^{W\times H \times C}=W_{d,t}^{1\times1\times C}X_{d,t}^{W\times H \times C_{1}}+b_{d,t}^{1\times1\times C},
\end{equation}
where $I_{t}$ is the corresponding instance representation, $W_{d,t}$ is the weight matrix of this $1\times1$ convolutional layer, $b_{d,t}$ is the bias matrix of this convolutional layer and $t=0, 1, 2,\cdots, T$.

Regarding the channel number, assume there are overall $C$ bag categories, then the instance representation $I_{t}$ also has $C$ channels so that the feature map of each channel corresponds to the response on a specific bag category, as it has been suggested in Eq.~\ref{eq2}. Thus, the number of $1\times1$ convolution filters in this layer is also $C$.

\begin{figure}[!t]
    \centering %插入的图片居中表示
	\includegraphics[width=3.2in]{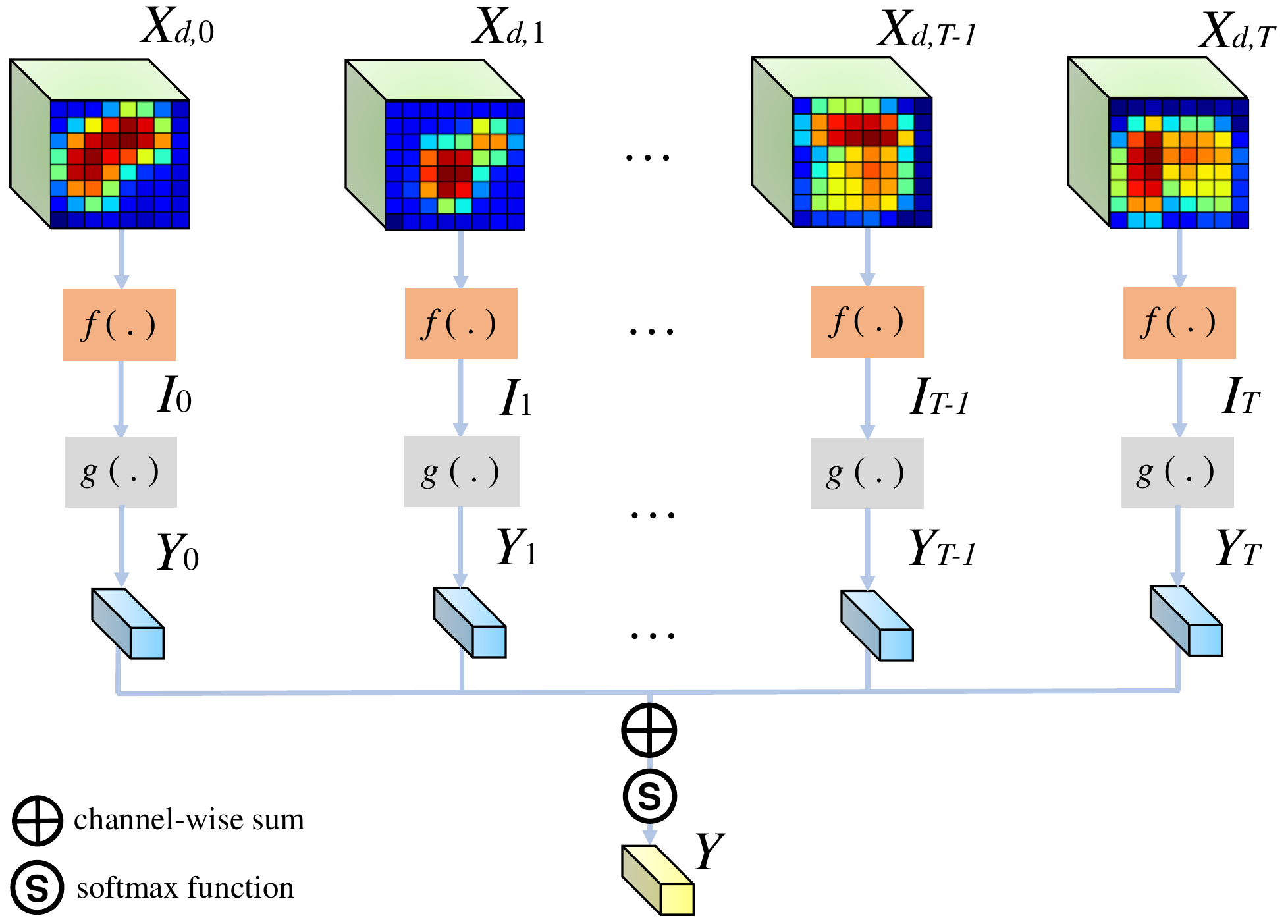} %插入的图，包括JPG,PNG,PDF,EPS等，放在源文件目录下
	\caption{Illustration on the instance representation and the generation of bag probability distribution.}  %图片的名称
	\label{fig5}   %标签，用作引用
\end{figure}

Apparently, each $1\times1$ image patch on the $W\times H$ sized feature map corresponds to an instance. As there are $C$ bag categories and the instance representation also has $C$ channels, each instance corresponds to a $C$-dimensional feature vector and thus each dimension corresponds to the feature response on the specific bag category (demonstrated in Fig.~\ref{fig5}).

\textit{3) Multi-grain instance representation:} After processed by Eq.~\ref{eq13}, each differential dilated convolutional feature representation $I_{t}$ generates an instance representation at the corresponding grain. Generally, the set of multi-grain instance representation $\{I_{t}\}$ can be presented as $\{I_{0}, I_{1}, \cdots, I_{T}\}$.

\textit{4) MIL aggregation function:} As is presented in Eq.~\ref{eq4}, under the instance space paradigm, the MIL aggregation function $g$ converts the instance representation directly into the bag probability distribution. On the other hand, the MIL aggregation function is required to be permutation invariant \cite{Dietterich1997Solving,Maron1998Multiple} so that the bag scheme prediction is invariant to the change of instance positions. Therefore, we utilize the mean based MIL pooling for aggregation.

Specifically, for the instance representation $I_{t}$ from each scale, assume each instance can be presented as $I_{t}^{w,h}$, where $1 \leq w \leq W$ and $1 \leq h \leq H$. Then, the generation of the bag probability distribution $Y_{t}$ from this grain is presented as
\begin{equation} \label{eq14}
Y_{t}=\frac{\sum_{w=1}^{W}\sum_{h=1}^{H} I_{t}^{w,h}}{W \times H},
\end{equation}

Apparently, after aggregation, $Y_{t}$ can be regarded as a $C$ dimensional feature vector. This process can be technically solved by a global average pooling (GAP) function in existing deep learning frameworks. 

\textit{5) Bag probability generation:} The final bag probability distribution $Y$ is the sum of the predictions from each grain, which is calculated as
\begin{equation} \label{eq15}
Y=softmax(\sum_{t=0}^{T}Y_{i}),
\end{equation}
where $softmax$ is the softmax function for normalization. 

To sum up, the pseudo code of all the above steps on learning multi-branch multi-instance representation is summarized in Algorithm~\ref{alg1}, in which $conv1d$ refers to the $1\times1$ convolution layer in Eq.~\ref{eq12}. 
%\vspace{-1.0em}

\begin{algorithm}[!t]  
  \caption{Learning Multi-branch Multi-instance Representation}  
  \label{alg1}  
  \begin{algorithmic}[1]  
    \Require  
    convolutional feature $X_{1}$, grain number $T$
    \Ensure  
    bag probability distribution $Y$, instance representation set $\{I_{t}\}$
      \State zero initialization $Y$
      
      %\State \% get multi-scale dilated convolution feature
      \For{$t=0$ $\rightarrow$ $T$}
      \State $X_{t}^{t} \leftarrow D_{t}(X_{1})$
      \EndFor
      
      \For{$t=0$ $\rightarrow$ $T$}
      \If{t $\geq$ 1}
      \State $X_{d,t} \leftarrow \|X_{1}^{t}-X_{1}^{t-1}\|$
      \Else
      \State \% $conv1d$: the convolutional layer in Eq.~\ref{eq12} 
      \State $X_{d,t} \leftarrow conv1d(X_{1})$
      \EndIf
      \EndFor
      
      %\State \% get the instance representation via transformation $f$
      \For{$t=0$ $\rightarrow$ $T$}
      \State $I_{t} \leftarrow f_{t}(X_{d,t})$
      %\State $I += I_{i} $
      \EndFor 
      
      %\State \% get the bag probability distribution via MIL aggregation function $g$
      \For {$t=0$ $\rightarrow$ $T$}  
      \State $Y_{t} \leftarrow g(I_{t})$
      \State $Y+=Y_{t}$;
      \EndFor
      %\State \% softmax function on $Y$ for normalization 
      \State $Y \leftarrow softmax(Y)$
  \end{algorithmic}  
\end{algorithm}

\subsection{Self-aligned Semantic Fusion Module}
\label{sec3.5}
\begin{figure}[!t]
    \centering %插入的图片居中表示
	\includegraphics[width=3.1in]{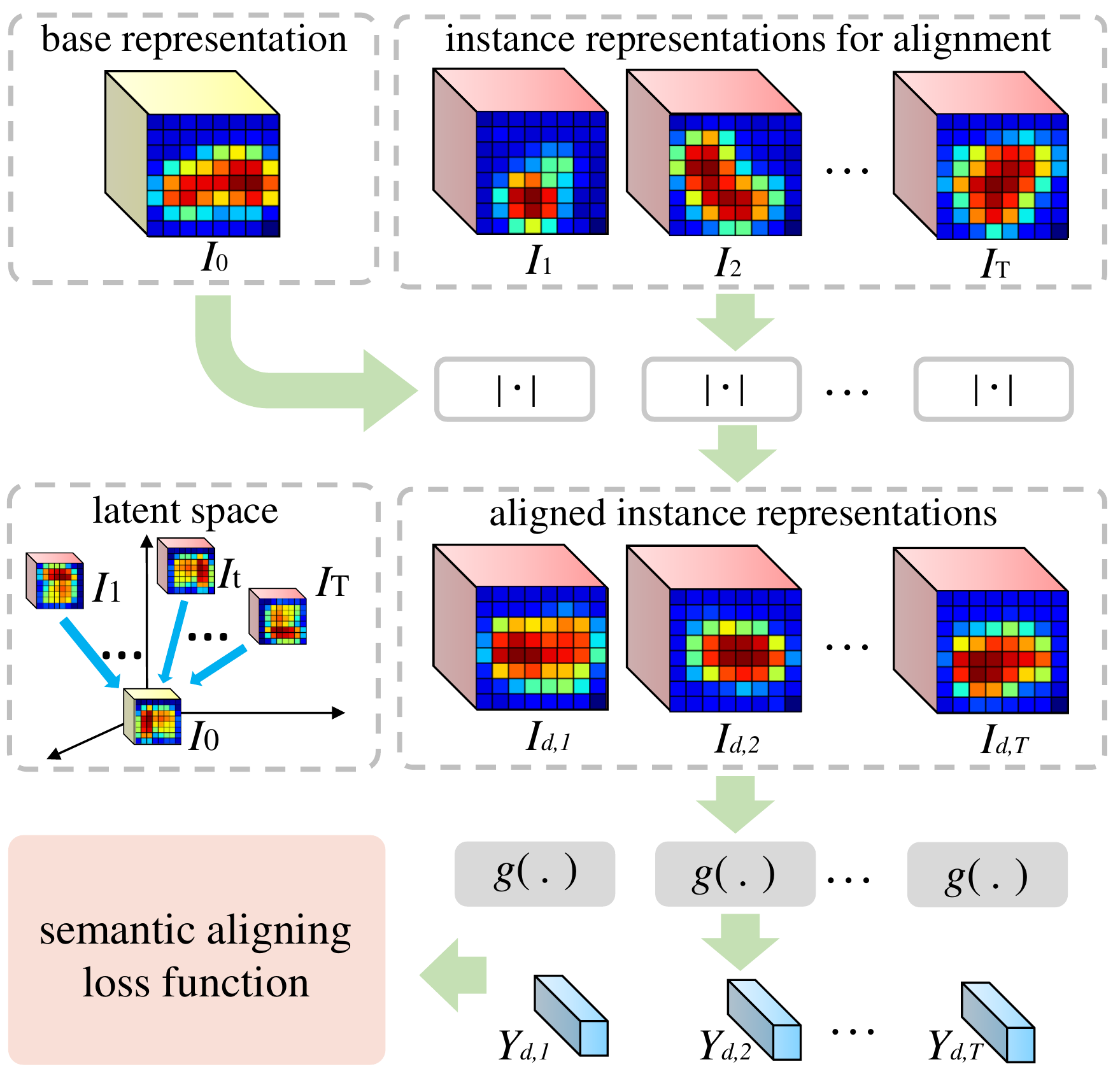} %插入的图，包括JPG,PNG,PDF,EPS等，放在源文件目录下
	\caption{Demonstration of our self-aligned semantic fusion (SSF) module.}  %图片的名称
	\label{fig6}   %标签，用作引用
\end{figure}

\textit{1) Motivation:} To make the instance representation  from different grains focus on the same bag scheme, we propose a bag scheme self-alignment strategy. Specifically, it at first finds the difference between a base instance representation and the instance representations from other grains, and then minimizes this difference by our semantic aligning loss function. Fig.~\ref{fig6}~offers an intuitive illustration of this module.

\textit{2) Base representation:} The instance representation $I_{0}$, only processed by a $1\times1$ convolutional layer rather than any dilated convolution, is selected as our base representation. One of the major reasons for using $I_{0}$ as the base representation is that the processing of the $1\times1$ convolutional layer can highlight the key local regions of an aerial scene.

\textit{3) Difference from base representation:} The absolute difference between other instance representation $I_{t}$ (here $t=1,2,\cdots,T$) and the base representation $I_{0}$ is calculated to depict the differences between the base representation and the other instance representation from different grains $t$. This process can be presented as 
\begin{equation} \label{eq16}
\begin{aligned}
%& I_{d,1}=\|I_{1}-I_{0}\|_{1}, \\
%& I_{d,2}=\|I_{2}-I_{0}\|_{1}, \\
%& \quad \quad \cdots \\
%& 
I_{d,t}=\|I_{t}-I_{0}\|, \\
%& \quad \quad \cdots \\
%& I_{d,T}=\|I_{T}-I_{0}\|_{1}, \\
%\setlength{\belowdisplayskip}{1pt}
\end{aligned}
\end{equation}
where $\|\cdot\|$ denotes the absolute difference, $I_{d,t}$ denotes the difference of each two instance representations at the corresponding grains, and $t=1, 2,\cdots, T$.

\textit{4) Bag scheme alignment:} By implementing the MIL aggregation function $g$ on $I_{d,t}$, the bag probability $Y_{d,t}$, depicting the difference of instance representations from adjacent grains, is generated. This process can be presented as
\begin{equation} \label{eq17}
%\begin{aligned}
Y_{d,t}=\frac{\sum_{w=1}^{W}\sum_{h=1}^{H} I_{d,t}^{w,h}}{W\times H},
%\setlength{\belowdisplayskip}{1pt}
%\end{aligned}
\end{equation}
where all the notations follow the paradigm in Eq.~\ref{eq14}, that is, $1 \leq w \leq W$ and $1 \leq h \leq H$, $W$ and $H$ denotes the width and height respectively. 

The overall bag scheme probability distribution differences $Y_{d}$ between the base instance representation $I_{d,0}$ and other instance representations $I_{d,t}$ (where $t=1, 2, \cdots, T$) can be calculated as
\begin{equation} \label{eq18}
\begin{aligned}
%Y_{d,t}=\frac{\sum_{w=1}^{W}\sum_{h=1}^{H} I_{d,t}^{w,h}}{WH},
Y_{d} &=softmax(\sum_{t=1}^{T} Y_{d,t}),\\
&=softmax(\frac{\sum_{t=1}^{T} \sum_{w=1}^{W}\sum_{h=1}^{H} I_{d,t}^{w,h}}{W\times H}),
\end{aligned}
\end{equation}
where $softmax$ denotes the softmax function. 

By minimizing the overall bag scheme probability differences $Y_{d}$, the bag prediction from each grain tends to be aligned to the same category. Technically, this minimization process is realized by our loss function in the next subsection. 

\begin{algorithm}[!t]  
  \caption{Bag Scheme Self-alignment Strategy}  
  \label{alg2}  
  \begin{algorithmic}[1]  
    \Require  
    instance representation set $\{I_{t}\}$, bag probability distribution $Y$, exact bag scheme $Y_{c}$
    \Ensure  
    loss function $L$ for optimization
      \State zero initialization $Y_{d}$
      
      %\State \% get the differences of each instance representation and the base representation $I_{0}$
      \For{$t=1$ $\rightarrow$ $T$}
      \State $I_{d,t} \leftarrow \|I_{t}-I_{0}\| $
      \EndFor

      %\State \% $g$ is our MIL aggregation function
      \For{$t=1$ $\rightarrow$ $T$}
      \State $Y_{d,t} \leftarrow g(I_{d,t})$
      \State $Y_{d}+=Y_{d,t}$
      %\State $I += I_{i} $
      \EndFor 
      
      %\State \% get the $L_{cls}$ and $L_{sealig}$ term, where $L_{crs}$ denotes the cross-entropy loss function form 
      \State $L_{cls} \leftarrow L_{crs}(Y,Y_{c})$
      \State $L_{sealig} \leftarrow L_{crs}(Y_{d},Y_{c})$
      
      %\State \% the overall loss, where $\alpha$ is a hyper-parameter 
      \State $L \leftarrow L_{cls}+ \alpha L_{sealig}$
  \end{algorithmic}  
\end{algorithm}

\subsection{Loss function}
\label{sec3.6}
\textit{1) Cross-entropy loss function:} Following the above notations, still assume $Y$ is the predicted bag probability distribution (in Eq.~\ref{eq15}), $Y_{c}$ is the exact bag category and there are overall $C$ categories. Then, the classic cross-entropy loss function serves as the classification loss $L_{cls}$, presented as
\begin{equation} \label{eq19}
\begin{aligned}
L_{cls}=-\frac{1}{C}\sum_{i=1}^C(Y_{c} \log Y_{i}+(1-Y_{c})\log (1-Y_{i})).
\end{aligned}
\end{equation}

\textit{2) Semantic-aligning loss function:} The formulation of the classic cross-entropy loss is also adapted to minimize the overall bag probability differences $Y_{d}$ in Eq.~\ref{eq18}. Thus, this semantic-aligning loss term $L_{sealig}$ is presented as
\begin{equation} \label{eq20}
\begin{aligned}
L_{sealig}=-\frac{1}{C}\sum_{i=1}^C(Y_{c} \log Y_{d,i}+(1-Y_{c})\log (1-Y_{d,i})).
\end{aligned}
\end{equation}

\textit{3) Overall loss:} The overall loss function $L$ to optimize the entire framework is the weighted average of the above two terms $L_{cls}$ and $L_{sealig}$, calculated as
\begin{equation} \label{eq21}
\begin{aligned}
L_= L_{cls} + \alpha L_{sealig},
\end{aligned}
\end{equation}
where $\alpha$ is the hyper-parameter to balance the impact of the above two terms. Empirically, we set $\alpha=5\times10{-4}$.

The pseudo code of our proposed overall bag scheme self-alignment strategy is provided in Algorithm~\ref{alg2}, which covers the content in our subsection~\ref{sec3.5}~and~\ref{sec3.6}.

\section{Experiment and Analysis}
\label{sec4}
\subsection{Datasets}

\subsubsection{UC Merced Land Use Dataset (UCM)} Till now, it is the most commonly-used aerial scene classification dataset. It has 2,100 samples in total and there are 100 samples for each of the 21 scene categories \cite{Yi2013Geographic}. All these samples have the size of 256$\times$256 with a 0.3-meter spatial resolution. Moreover, all these samples are taken from the aerial craft, and both the illumination condition and the viewpoint of all these aerial scenes is quite close. 

\subsubsection{Aerial Image Dataset (AID)} It is a typical large-scale aerial scene classification benchmark with an image size of 600$\times$600 \cite{Xia2017AID}. It has 30 scene categories with a total amount of 10,000 samples. The sample number per class varies from 220 to 420. As the imaging sensors in photographing the aerial scenes are more varied in AID benchmark, the illumination conditions and viewpoint are also more varied. Moreover, the spatial resolution of these samples varies from 0.5 to 8 meters. 

\subsubsection{Northwestern Polytechnical University (NWPU) dataset} This benchmark is more challenging than the UCM and AID benchmarks as the spatial resolution of samples varies from 0.2 to 30 meters \cite{Gong2017Remote}. It has 45 scene categories and 700 samples per class. All the samples have a fixed image size of $256\times256$. Moreover, the imaging sensors and imaging conditions are more varied and complicated than AID.

\subsection{Evaluation protocols}
Following the existing experiment protocols \cite{Xia2017AID,Gong2017Remote}, we report the overall accuracy (OA) in the format of 'average$\pm$deviation' from ten independent runs on all these three benchmarks. 

Experiments on UCM, AID and NWPU dataset are all in accordance with the corresponding training ratio settings. To be specific, for UCM the training set proportions are 50\% and 80\% respectively, for AID the training set proportions are 20\% and 50\% respectively, and for NWPU the training set proportions are 10\% and 20\% respectively.  

\begin{table}[!t]  
    \centering
    \caption{Data partition and evaluation protocols of the three aerial scene classification benchmarks following the evaluation protocols \cite{Xia2017AID,Gong2017Remote}, where \textit{runs} denotes the required independent repetitions to report the classification accuracy. }
    \begin{tabular}{ccccc} 
    \hline
    & \#classes & \tabincell{c}{\# samples} & \tabincell{c}{training / testing ratio} & runs\\ 
    \hline
    UCM & 21 & 210 & 50\% / 50\%, 80\% / 20\% \cite{Xia2017AID} & 10 \\
    %\hline
    AID & 30 & 220-420 & 20\% / 80\%, 50\% / 50\% \cite{Xia2017AID} & 10 \\
    %\hline
    NWPU &	45 & 700 & 10\% / 90\%, 20\% / 80\%\cite{Gong2017Remote} & 10\\
    \hline
    \end{tabular} 
    \label{tab_data}
\end{table}

\subsection{Experimental Setup}

\textit {Parameter settings:}
In our AGOS, $C_{1}$ is set 256, indicating there are 256 channels for each dilated convolutional filter. Moreover, $T$ is set 3, which means there are 4 branches in our AGOS module. Finally, $C$ is set 21, 30 and 45 respectively when trained on UCM, AID and NWPU benchmark respectively, which equals to the total scene category number of these three benchmarks. 

\textit {Model initialization:}
A set of backbones, including ResNet-50, ResNet-101 and DenseNet-121, all utilize pre-trained parameters on ImageNet as the initial parameters. For the rest of our AGOS framework, we use random initialization for weight parameters with a standard deviation of 0.001. All bias parameters are set zero for initialization.

\textit {Training procedure:}
The model is optimized by the Adam optimizer with $\beta_{1}=0.9$ and $\beta_{2}=0.999$. Moreover, the batch size is set 32. 
The initial learning rate is set to be 0.0001 and is divided by 0.5 every 30 epochs until finishing 120 epochs. To avoid the potential over-fitting problem, $L_2$ normalization with a parameter setting of $5\times10^{-4}$ is utilized and a dropout rate of 0.2 is set in all the experiments.

\textit {Other implementation details:}
Our experiments were conducted under the TensorFlow deep learning framework by using the Python program language. All the experiments were implemented on a work station with 64 GB RAM and a i7-10700 CPU. Moreover, two RTX 2080 SUPER GPUs are utilized for acceleration. Our source code is available at \href{https://github.com/BiQiWHU/AGOS}{https://github.com/BiQiWHU/AGOS}.

\subsection{Comparison with state-of-the-art approaches}
We compare the performance of our AGOS with three hand-crafted features (PLSA, BOW, LDA) \cite{Xia2017AID,Gong2017Remote}, three typical CNN models (AlexNet, VGG, GoogLeNet) \cite{Xia2017AID,Gong2017Remote}, seventeen latest CNN-based state-of-the-art approaches (MIDCNet \cite{Bi2019Multiple}, RANet \cite{Bi2020RADC}, APNet \cite{Bi2019APDC}, SPPNet \cite{Han2017Pre}, DCNN \cite{Gong2018When}, TEXNet \cite{Rao2017Binary}, MSCP \cite{He2018Remote}, VGG+FV \cite{Li2017Integrating}, DSENet \cite{Wang2021Enhanced}, MS2AP \cite{Bi2021MS}, MSDFF \cite{Xue2020Multi}, CADNet \cite{Tong2020Channel}, LSENet \cite{Bi2021LSENet}, GBNet \cite{Sun2020GBN}, MBLANet \cite{Chen2021MBLANet}, MG-CAP \cite{Wang2020MGCAP}, Contourlet CNN \cite{Liu2021CCNN}), one RNN-based approach (ARCNet \cite{WangQi2018RA}), two auto-encoder based approaches (SGUFL \cite{Zhang2014Saliency}, PARTLETS \cite{Cheng2015Effective}) and two GAN-based approaches (MARTA \cite{Lin2017MARTA}, AGAN \cite{Yu2019AGAN}) respectively. The performance under the backbone of ResNet-50, ResNet-101 and DenseNet-121 is all reported for fair evaluation as some latest methods \cite{Xue2020Multi,Tong2020Channel} use much deeper networks as backbone.

\subsubsection{Results and comparison on UCM}
In Table~\ref{tab1}, the classification accuracy of our AGOS and other state-of-the-art approaches is listed. It can be seen that:

\begin{table}[!t]  
    \centering
    \begin{threeparttable}
    \caption{Classification accuracy of our AGOS and other approaches on UCM dataset. Results presented in the form of 'average$\pm$deviation'\cite{Xia2017AID}; Metrics presented in \%; $H$, $C$, $R$, $A$ and $G$ denote hand-crafted, CNN, RNN, Auto-encoder and GAN based approaches respectively. In bold and in blue denotes the best and second best results.}
    \begin{tabular}{cccc} 
    \hline
    \multirow{2}{*}{Method} & \multirow{2}{*}{Type \& Year} & \multicolumn{2}{c}{Training ratio} \\ 
    \cline{3-4}
    ~ & ~ & 50\% & 80\% \\
    \hline
    PLSA(SIFT) \cite{Xia2017AID} & $H$, 2017 & 67.55$\pm$1.11 & 71.38$\pm$1.77\\
    %\hline
    BoVW(SIFT) \cite{Xia2017AID} & $H$, 2017 & 73.48$\pm$1.39 & 75.52$\pm$2.13\\
    %\hline
    LDA(SIFT) \cite{Xia2017AID} & $H$, 2017 & 59.24$\pm$1.66 & 75.98$\pm$1.60\\
    \hline
    AlexNet \cite{Xia2017AID} & $C$, 2017 & 93.98$\pm$0.67 & 95.02$\pm$0.81\\
    VGGNet-16 \cite{Xia2017AID} & $C$, 2017 & 94.14$\pm$0.69 & 95.21$\pm$1.20\\
    GoogLeNet \cite{Xia2017AID} & $C$, 2017 & 92.70$\pm$0.60 & 94.31$\pm$0.89\\
    \hline
    ARCNet \cite{WangQi2018RA} & $R$, 2018 & 96.81$\pm$0.14 & 99.12$\pm$0.40\\
    \hline
    SGUFL \cite{Zhang2014Saliency} & $A$, 2014 & ---- & 82.72$\pm$1.18 \\
    PARTLETS \cite{Cheng2015Effective} & $A$, 2015 & 88.76$\pm$0.79 & ----\\
    \hline
    MARTA \cite{Lin2017MARTA} & $G$, 2017 & 85.50$\pm$0.69 & 94.86$\pm$0.80 \\
    AGAN \cite{Yu2019AGAN} & $G$, 2019 & 89.06$\pm$0.50 & 97.69$\pm$0.69\\
    \hline
    VGG+FV \cite{Li2017Integrating} & $C$, 2017 & ---- & 98.57$\pm$0.34\\
    SPPNet \cite{Han2017Pre} & $C$, 2017 & 94.77$\pm$0.46* & 96.67$\pm$0.94*\\
    TEXNet \cite{Rao2017Binary} & $C$, 2017 & 94.22$\pm$0.50 & 95.31$\pm$0.69\\
    DCNN \cite{Gong2018When} & $C$, 2018 & ---- & 98.93$\pm$0.10\\
    MSCP \cite{He2018Remote} & $C$, 2018 & ---- & 98.36$\pm$0.58\\
    APNet \cite{Bi2019APDC} & $C$, 2019 & 95.01$\pm$0.43 & 97.05$\pm$0.43\\
    MSDFF \cite{Xue2020Multi} & $C$, 2020 & 98.85$\pm$---- &  99.76$\pm$----  \\
    CADNet \cite{Tong2020Channel} & $C$, 2020 & 98.57$\pm$0.33 &  99.67$\pm$0.27  \\
    MIDCNet \cite{Bi2019Multiple} & $C$, 2020 & 94.93$\pm$0.51 & 97.00$\pm$0.49\\
    RANet \cite{Bi2020RADC} & $C$, 2020 & 94.79$\pm$0.42 & 97.05$\pm$0.48\\
    GBNet \cite{Sun2020GBN} & $C$, 2020 & 97.05$\pm$0.19 &  98.57$\pm$0.48 \\
    MG-CAP \cite{Wang2020MGCAP} & $C$, 2020 & ---- & 99.00$\pm$0.10 \\
    DSENet \cite{Wang2021Enhanced} & $C$, 2021 & 96.19$\pm$0.13 & 99.14 $\pm$0.22  \\
    MS2AP \cite{Bi2021MS} & $C$, 2021 & 98.38$\pm$0.35 & 99.01 $\pm$0.42  \\
    Contourlet CNN \cite{Liu2021CCNN} & $C$, 2021 & ---- & 98.97$\pm$0.21  \\
    LSENet \cite{Bi2021LSENet} & $C$, 2021 & 97.94$\pm$0.35 & 98.69$\pm$0.53 \\
    MBLANet \cite{Chen2021MBLANet} & $C$, 2021 & ---- & 99.64$\pm$0.12 \\
    DMSMIL \cite{ICASSP2021} & $C$, 2021 & 99.09$\pm$0.36 & 99.45$\pm$0.32\\
    \hline
    \textbf{AGOS} (ResNet-50) & $C$ & 99.24$\pm$0.22 & 99.71$\pm$0.25 \\
    \textbf{AGOS} (ResNet-101) & $C$ & \textcolor{blue}{99.29$\pm$0.23} & \textcolor{blue}{99.86$\pm$0.17}\\
    \textbf{AGOS} (DenseNet-121) & $C$ & \textbf{99.34$\pm$0.20} & \textbf{99.88$\pm$0.13}\\
    \hline
    \end{tabular} 
    \begin{tablenotes}
        \footnotesize
        \item ‘----’: not reported, ‘*’: not reported \& conducted by us
    \end{tablenotes}
    \label{tab1}
    \end{threeparttable}
\end{table}

\begin{figure*}[!t]   
    \begin{tabular}{cc} 
    \subfigure[UCM 50\%]{ 
    \begin{minipage}[t]{0.31\textwidth} 
    \centering 
    \includegraphics[width=2.2in,height=1.5in]{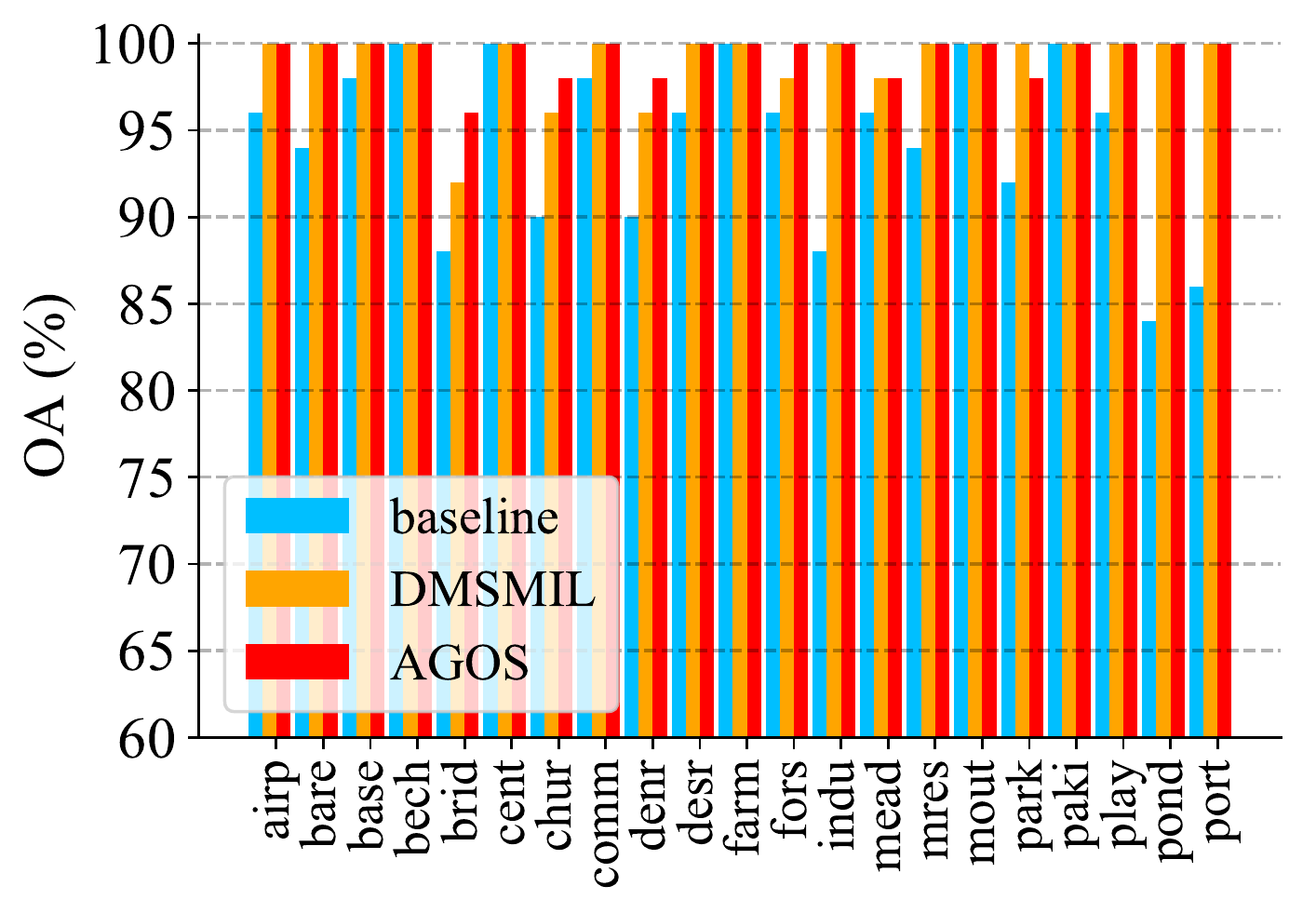}
    \end{minipage}}     
    %\hspace{0.01\textwidth} 
    \subfigure[UCM 80\%]{ 
    \begin{minipage}[t]{0.31\textwidth} 
    \centering 
    \includegraphics[width=2.2in,height=1.5in]{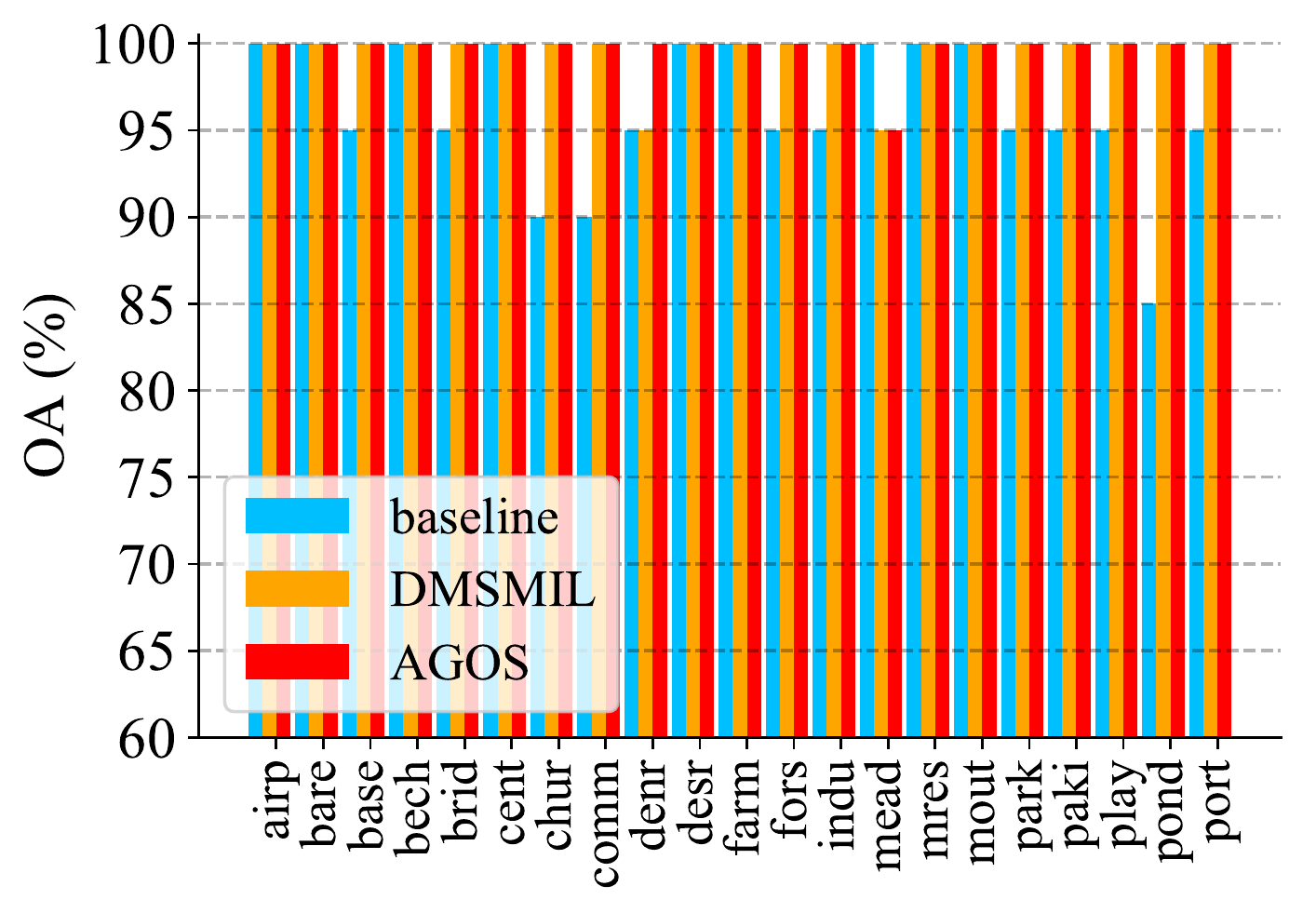}
    \end{minipage}}
       %\\
    %\hspace{0.01\textwidth} 
    \subfigure[AID 20\%]{ 
    \begin{minipage}[t]{0.31\textwidth} 
    \centering 
    \includegraphics[width=2.2in,height=1.5in]{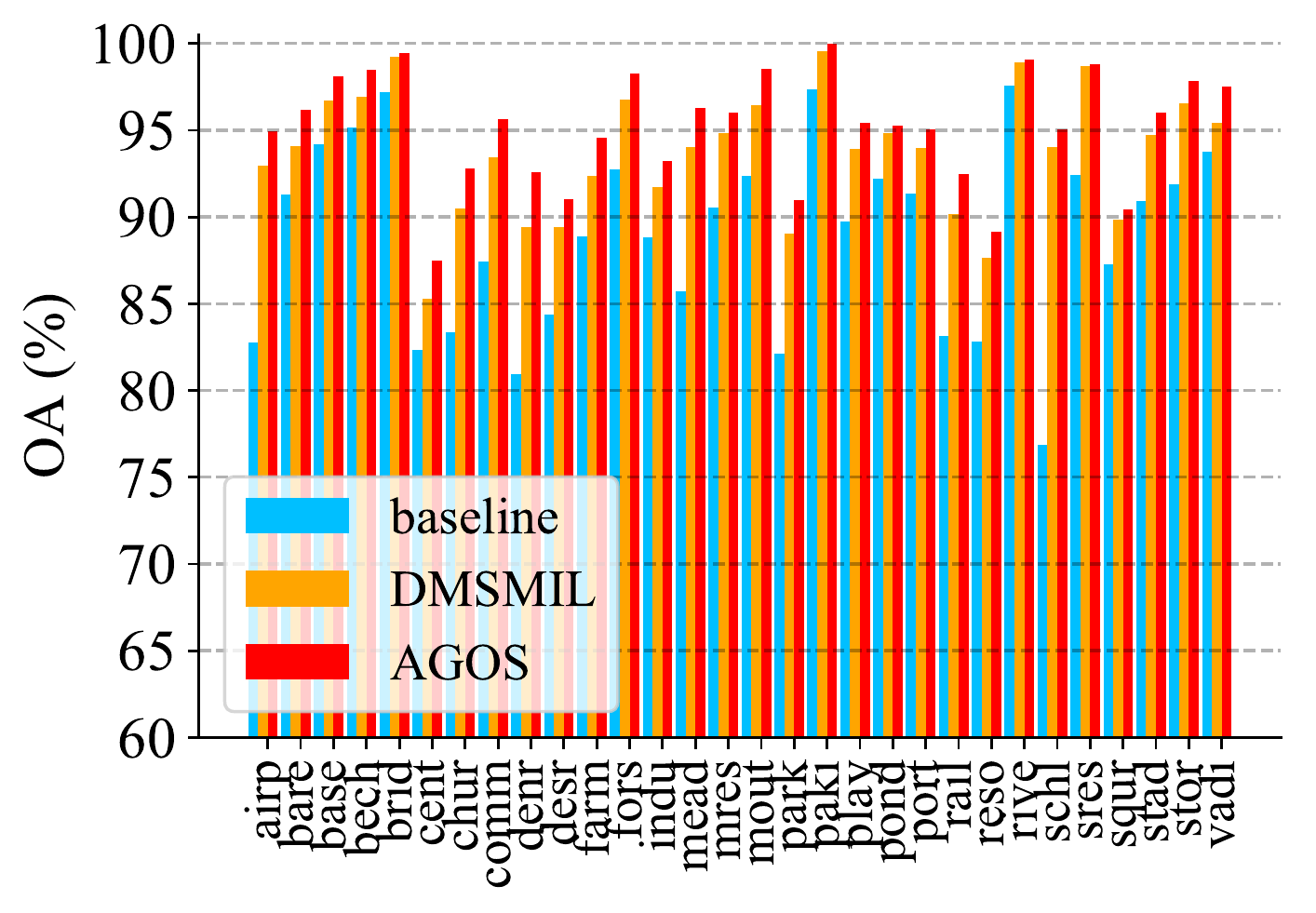}
    \end{minipage}}
       \\
    \subfigure[AID 50\%]{ 
    \begin{minipage}[t]{0.31\textwidth} 
    \centering 
    \includegraphics[width=2.2in,height=1.5in]{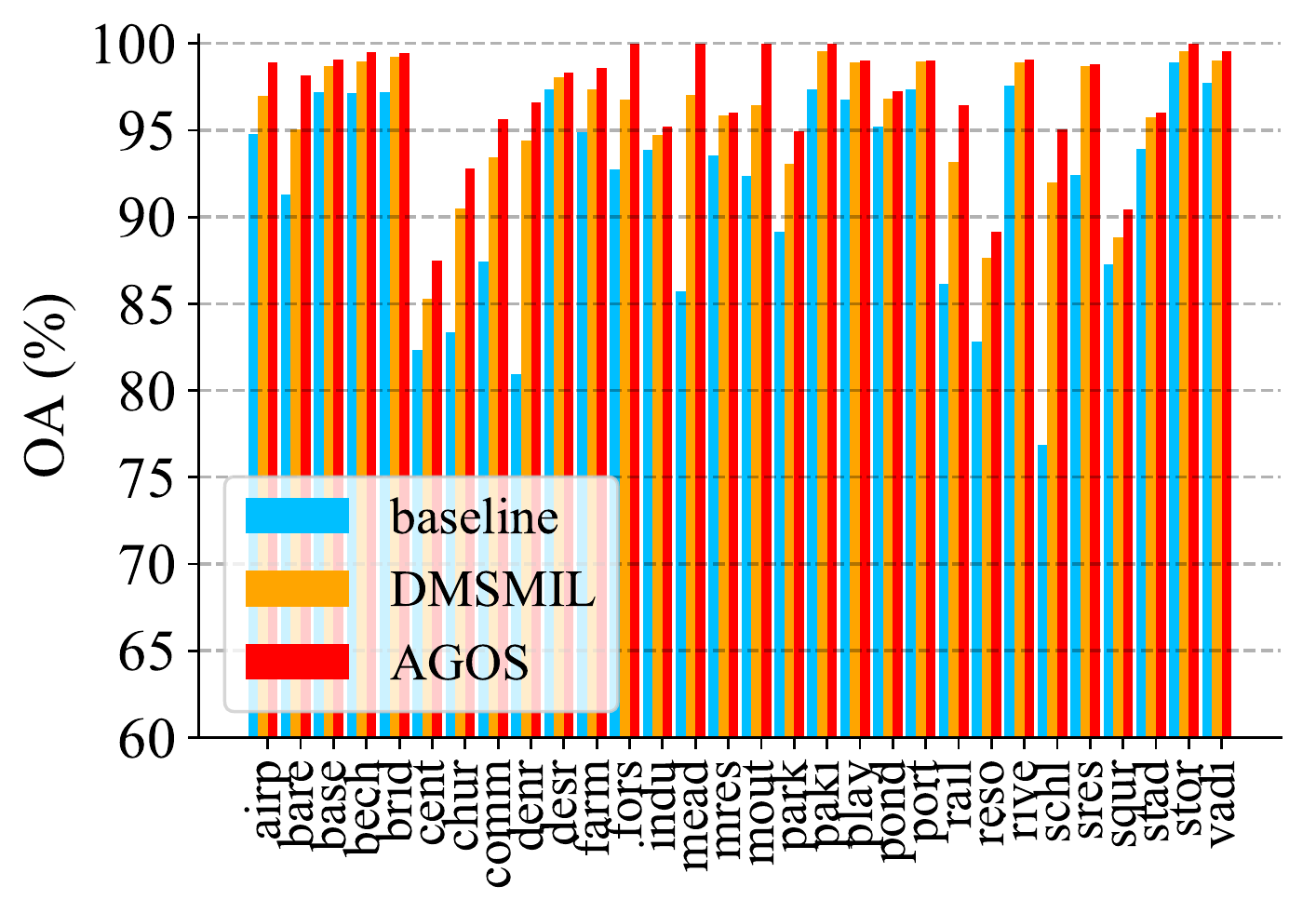}
    \end{minipage}}  
         %\\
    %\hspace{0.01\textwidth} 
    \subfigure[NWPU 10\%]{ 
    \begin{minipage}[t]{0.31\textwidth} 
    \centering 
    \includegraphics[width=2.2in,height=1.5in]{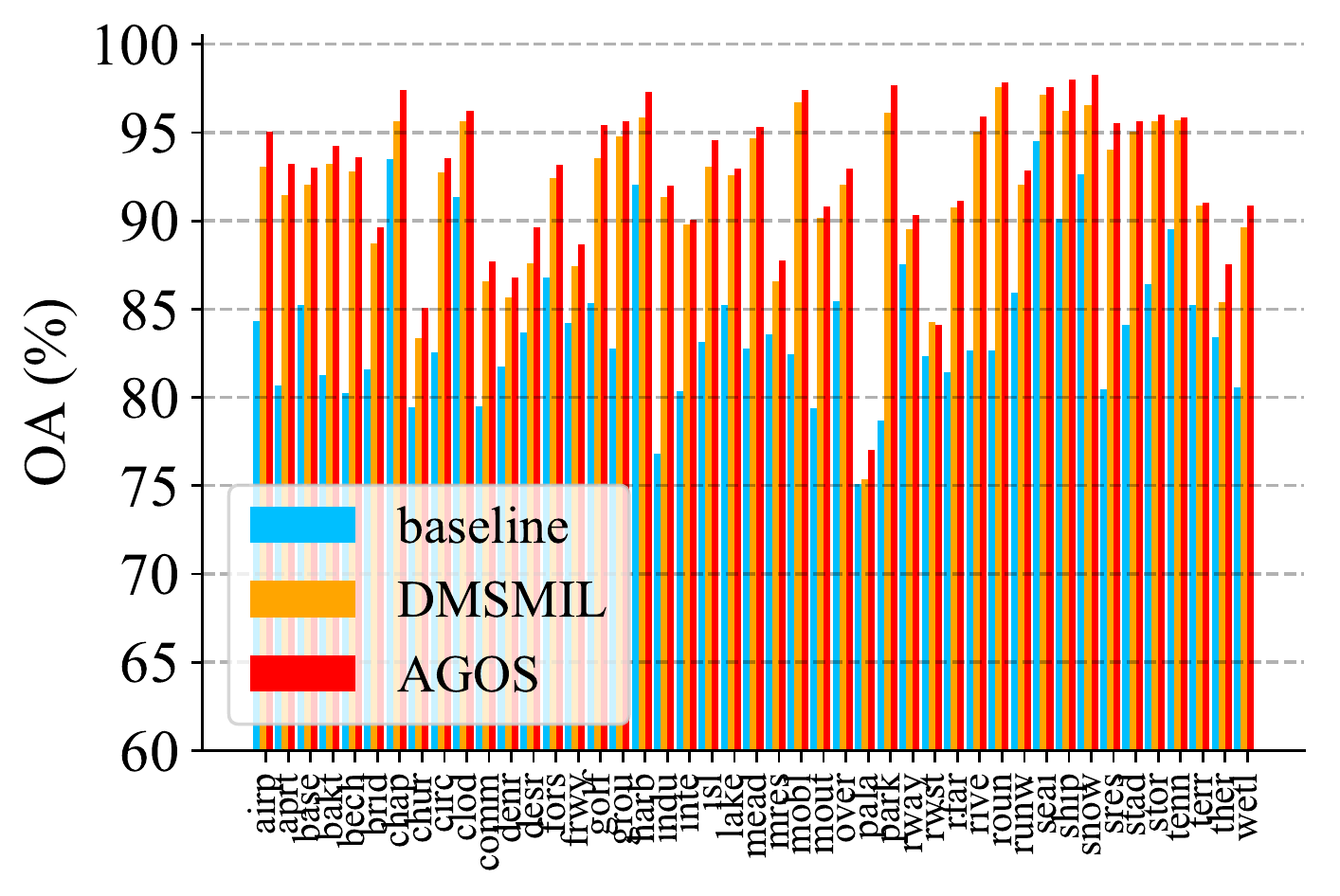} 
    \end{minipage}} 
   % \hspace{0.01\textwidth} 
    \subfigure[NWPU 20\%]{ 
    \begin{minipage}[t]{0.31\textwidth} 
    \centering 
    \includegraphics[width=2.2in,height=1.5in]{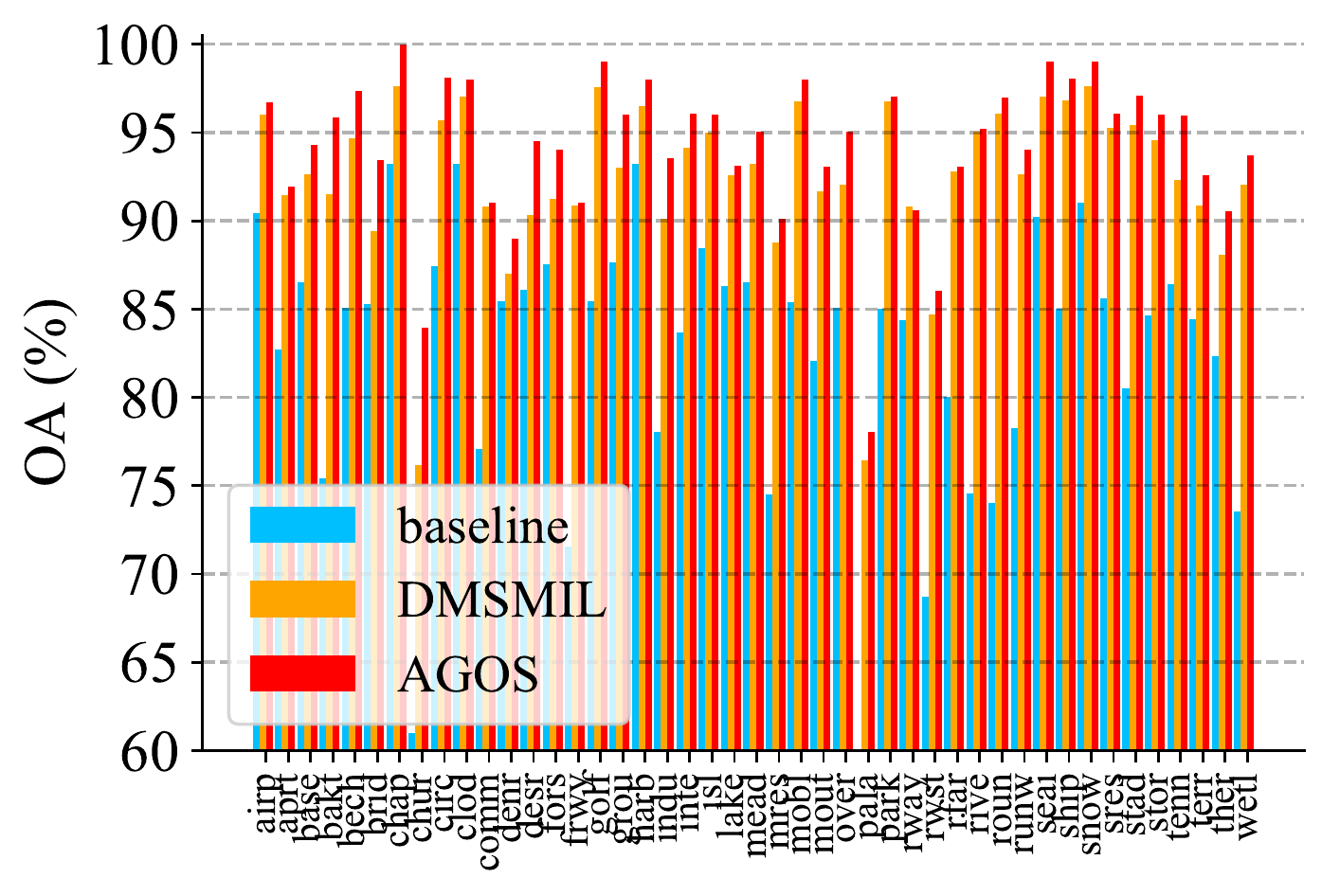} 
    \end{minipage}}  \\
    \end{tabular}   
\caption{Classification accuracy of the ResNet-50 baseline and our proposed AGOS with ResNet-50 backbone on three datasets. \textit{DMSMIL} with orange bar denotes the performance of our initial version \cite{ICASSP2021}; \textit{AGOS} with red bar denotes the performance of our current version.}   
\label{fig7}   
\end{figure*}

\begin{enumerate}[(1)]
\item Both our current AGOS framework and its initial version \cite{ICASSP2021} outperform the existing state-of-the-art approaches on both cases when the training ratios are 50\% and 80\% respectively, including CNN based, RNN based, auto-encoder based and GAN based approaches. Plus, no matter using lighter backbone ResNet-50 or stronger backbone ResNet-101 and DenseNet-121, our AGOS shows superior performance against all the compared methods.
\item Our AGOS framework significantly outperforms the existing approaches that exploit the multi-scale representation for aerial scenes \cite{Han2017Pre,Bi2021MS,He2018Remote,Li2017Integrating}.
\item Generally speaking, other approaches that achieve the most competitive performance usually highlight the key local regions of an aerial scene \cite{WangQi2018RA,Bi2019Multiple,Bi2020RADC,He2018Remote}. For the auto-encoder and GAN based approaches, as this aspect remains unexplored, their performance is relatively weak. 
\end{enumerate}

Per-category classification accuracy (with ResNet-50 backbone) when the training ratios are 50\% and 80\% is displayed in Fig.~\ref{fig7}~(a), (b) respectively. It is observed that almost all the samples in the UCM are correctly classified. Still, it is notable that the hard-to-distinguish scene categories such as dense residential, medium residential and sparse residential are all identified correctly.

The potential explanations are summarized as follows. 

\begin{enumerate}[(1)]
\item Compared with ground images, aerial images are usually large-scale. Thus, the highlight of key local regions related to the scene scheme is vital. The strongest-performed approaches, both CNN based  \cite{WangQi2018RA,Bi2019Multiple,Bi2020RADC,He2018Remote,Gong2018When} and our AGOS, take the advantage of these strategies.
\item Another important aspect for aerial scene classification is to consider the case that the sizes of key objects in aerial scenes vary a lot. Hence, it is observed that many competitive approaches are utilizing the multi-scale feature representation \cite{Han2017Pre,Bi2021MS,He2018Remote,Li2017Integrating}. Our AGOS also takes advantage of this and contains a multi-grain perception module. More importantly, our AGOS further allows the instance representation from each grain to focus on the same scene scheme, and thus the performance improves.
\item Generally speaking, the performance of auto-encoder \cite{Zhang2014Saliency,Cheng2015Effective} and GAN \cite{Lin2017MARTA,Yu2019AGAN} based solutions is not satisfactory, which may also be explained from the lack of the above capabilities such as the highlight of key local regions and multi-grain representation.
\end{enumerate}

\subsubsection{Results and comparison on AID}
In Table~\ref{tab2}, the results of our AGOS and other state-of-the-art approaches on AID are listed. Several observations can be made.

\begin{table}[!t]  
    \centering
    \begin{threeparttable}
    \caption{Classification accuracy of our proposed AGOS and other approaches on AID dataset. Results presented in the form of 'average$\pm$deviation'\cite{Xia2017AID}; Metrics presented in \%; $H$, $C$, $R$ and $G$ denote hand-crafted, CNN, RNN and GAN based approaches. In bold and in blue denotes the best and second best results.}
    \begin{tabular}{cccc} 
    \hline
    \multirow{2}{*}{Method} & \multirow{2}{*}{Type} & \multicolumn{2}{c}{Training ratio} \\ 
    \cline{3-4}
    ~ & ~ & 20\% & 50\% \\
    \hline
    PLSA(SIFT) \cite{Xia2017AID} & $H$, 2017 & 56.24$\pm$0.58 & 63.07$\pm$1.77\\
    %\hline
    BoVW(SIFT) \cite{Xia2017AID} & $H$, 2017 & 62.49$\pm$0.53 & 68.37$\pm$0.40\\
    %\hline
    LDA(SIFT) \cite{Xia2017AID} & $H$, 2017 & 51.73$\pm$0.73 & 68.96$\pm$0.58\\
    \hline
    AlexNet \cite{Xia2017AID} & $C$, 2017 & 86.86$\pm$0.47 & 89.53$\pm$0.31\\
    VGGNet-16 \cite{Xia2017AID} & $C$, 2017 & 86.59$\pm$0.29 & 89.64$\pm$0.36\\
    GoogLeNet \cite{Xia2017AID} & $C$, 2017 & 83.44$\pm$0.40 & 86.39$\pm$0.55\\
    \hline
    ARCNet \cite{WangQi2018RA} & $R$, 2018 & 88.75$\pm$0.40 & 93.10$\pm$0.55\\
    \hline
    MARTA \cite{Lin2017MARTA} & $G$, 2017 & 75.39$\pm$0.49 & 81.57$\pm$0.33  \\
    AGAN \cite{Yu2019AGAN} & $G$, 2019 & 78.95$\pm$0.23 & 84.52$\pm$0.18\\
    \hline
    SPPNet \cite{Han2017Pre} & $C$, 2017 & 87.44$\pm$0.45* & 91.45$\pm$0.38*\\
    TEXNet \cite{Rao2017Binary} & $C$, 2017 & 87.32$\pm$0.37 & 90.00$\pm$0.33\\
    MSCP \cite{He2018Remote} & $C$, 2018 & 91.52$\pm$0.21 & 94.42$\pm$0.17\\
    APNet \cite{Bi2019APDC} & $C$, 2019 & 88.56$\pm$0.29 & 92.15$\pm$0.29\\
    MSDFF \cite{Xue2020Multi} & $C$, 2020 & 93.47$\pm$---- &  96.74$\pm$----  \\
    CADNet \cite{Tong2020Channel} & $C$, 2020 & \textcolor{blue}{95.73$\pm$0.22} &  97.16$\pm$0.26  \\
    MIDCNet \cite{Bi2019Multiple} & $C$, 2020 & 88.26$\pm$0.43 & 92.53$\pm$0.18\\
    RANet \cite{Bi2020RADC} & $C$, 2020 & 88.12$\pm$0.43 & 92.35$\pm$0.19  \\
    GBNet \cite{Sun2020GBN} & $C$, 2020 & 92.20$\pm$0.23 & 95.48$\pm$0.12 \\
    MG-CAP \cite{Wang2020MGCAP} & $C$, 2020 & 93.34$\pm$0.18 & 96.12$\pm$0.12 \\
    M$F^{2}$Net \cite{Xu2020MFNet} & $C$, 2020 & 93.82$\pm$0.26 & 95.93$\pm$0.23 \\
    DSENet \cite{Wang2021Enhanced} & $C$, 2021 & 94.02$\pm$0.21 & 94.50$\pm$0.30   \\
    MS2AP \cite{Bi2021MS} & $C$, 2021 & 92.19$\pm$0.22 & 94.82$\pm$0.20  \\
    Contourlet CNN \cite{Liu2021CCNN} & $C$, 2021 & ---- & 96.65$\pm$0.24 \\
    LSENet \cite{Bi2021LSENet} & $C$, 2021 & 94.07$\pm$0.19 & 95.82$\pm$0.19 \\
    MBLANet \cite{Chen2021MBLANet} & $C$, 2021 & 95.60$\pm$0.17 & 97.14$\pm$0.03 \\
    LiGNet \cite{Xu2021LiGNet} & $C$, 2021 & 94.17$\pm$0.25 & 96.19$\pm$0.28 \\
    DMSMIL \cite{ICASSP2021} & $C$, 2021 & 93.98$\pm$0.17 & 95.65$\pm$0.22\\
    \hline
    \textbf{AGOS} (ResNet-50) & $C$ & 94.99$\pm$0.24 & 97.01$\pm$0.18\\
    \textbf{AGOS} (ResNet-101) & $C$ & 95.54$\pm$0.23 & \textcolor{blue}{97.22$\pm$0.19}\\
    \textbf{AGOS} (DenseNet-121) & $C$ & \textbf{95.81$\pm$0.25} & \textbf{97.43$\pm$0.21}\\
    \hline
    \end{tabular} 
    \begin{tablenotes}
        \footnotesize
        \item ‘----’: not reported, ‘*’: not reported \& conducted by us
    \end{tablenotes}
    \label{tab2}
    \end{threeparttable}
\end{table}

\begin{enumerate}[(1)]
\item Our proposed AGOS with DenseNet-121 outperforms all the state-of-the-art methods under both the training ratio of 20\% and 50\%. Its ResNet-101 version achieves the second best results under training ratio 50\%. Moreover, AGOS with ResNet-50 and our former version \cite{ICASSP2021} also achieves a satisfactory performance on both experiments. 
\item Other state-of-the-art approaches that either highlight the key local regions \cite{Gong2018When,Bi2019Multiple,WangQi2018RA,Bi2020RADC} or build a multi-scale representation \cite{Han2017Pre,Bi2021MS,He2018Remote,Li2017Integrating} also perform well on both two experiments. 
\item Similar to the situations in UCM, the strongest performance mainly comes from CNN based methods \cite{Gong2018When,Bi2019Multiple,WangQi2018RA,Bi2020RADC}, while the performance of GAN based methods is far from satisfactory \cite{Lin2017MARTA,Yu2019AGAN}. 
\end{enumerate}
 
Per-category classification accuracy under the training ratio of 20\% and 50\% is shown in Fig.~\ref{fig7}~(c) and (d) respectively. It can be seen that most scene categories are well distinguished, and some categories difficult to classify, \textit{i.e.}, dense residential, medium residential and sparse residential, are also classified well by our solution. Possible explanations include: 
\begin{enumerate}[(1)]
\item The sample size in AID is generally larger than UCM, and the key objects to determine the scene category are more varied in terms of sizes. As our AGOS can highlight the key local regions via MIL and can build a more discriminative multi-grain representation than existing multi-scale aerial scene classification methods \cite{Han2017Pre,Bi2021MS,He2018Remote,Li2017Integrating}, it achieves the strongest performance. 
\item Highlighting the key local regions is also quite important to enhance the aerial scene representation capability for the deep learning frameworks \cite{Gong2018When,Bi2019Multiple,WangQi2018RA,Bi2020RADC}, and this can also be one of the major reasons to account for the weak performance of GAN based methods \cite{Lin2017MARTA,Yu2019AGAN}.
\item As there are much more training samples in AID benchmark than in UCM, the gap of representation capability between traditional hand-crafted features and deep learning based approaches becomes more obvious. In fact, it is a good example to illustrate that the traditional hand-crafted feature based methods are far from enough to depict the complexity of the aerial scenes. 
\end{enumerate}

\subsubsection{Results and comparison on NWPU}
Table~\ref{tab3} lists the per-category classification results of our AGOS and other state-of-the-art approaches on NWPU benchmark. Several observations similar to the AID can be made.

\begin{table}[!t]  
    \centering
    \begin{threeparttable}
    \caption{Overall accuracy of the proposed AGOS and other approaches on NWPU dataset. Results presented in the form of 'average$\pm$deviation'\cite{Gong2017Remote}; Metrics presented in \%; $H$, $C$, $R$ and $G$ denote hand-crafted, CNN, RNN and GAN based approaches respectively. In bold and in blue denotes the best and second best results.}
    \begin{tabular}{cccc} 
    \hline
    \multirow{2}{*}{Method} & \multirow{2}{*}{Type} & \multicolumn{2}{c}{Training ratio} \\ 
    \cline{3-4}
    ~ & ~ & 10\% & 20\% \\
    \hline
    BoVW(SIFT) \cite{Gong2017Remote} & $H$, 2017 & 41.72$\pm$0.21 & 44.97$\pm$0.28\\
    \hline
    AlexNet \cite{Gong2017Remote} & $C$, 2017 & 76.69$\pm$0.21 & 79.85$\pm$0.13\\
    VGGNet-16 \cite{Gong2017Remote} & $C$, 2017 & 76.47$\pm$0.18 & 79.79$\pm$0.15\\
    GoogLeNet \cite{Gong2017Remote} & $C$, 2017 & 76.19$\pm$0.38 & 78.48$\pm$0.26\\
    \hline
    MARTA \cite{Lin2017MARTA} & $G$, 2017 & 68.63$\pm$0.22 & 75.03$\pm$0.28  \\
    AGAN \cite{Yu2019AGAN} & $G$, 2019 & 72.21$\pm$0.21 & 77.99$\pm$0.19 \\
    \hline
    SPPNet \cite{Han2017Pre} & $C$, 2017 & 82.13$\pm$0.30* & 84.64$\pm$0.23*\\
    DCNN \cite{Gong2018When} & $C$, 2018 & 89.22$\pm$0.50 & 91.89$\pm$0.22\\
    MSCP \cite{He2018Remote} & $C$, 2018 & 85.33$\pm$0.17 & 88.93$\pm$0.14\\
    MSDFF \cite{Xue2020Multi} & $C$, 2020 & 91.56$\pm$---- &  93.55$\pm$----  \\
    CADNet \cite{Tong2020Channel} & $C$, 2020 & 92.70$\pm$0.32 &  94.58$\pm$0.26  \\
    MIDCNet \cite{Bi2019Multiple} & $C$, 2020 & 85.59$\pm$0.26 & 87.32$\pm$0.17\\
    RANet \cite{Bi2020RADC} & $C$, 2020 & 85.72$\pm$0.25 & 87.63$\pm$0.28 \\
    MG-CAP \cite{Wang2020MGCAP} & $C$, 2020 & 90.83$\pm$0.12 & 92.95$\pm$0.11 \\
    M$F^{2}$Net \cite{Xu2020MFNet} & $C$, 2020 & 90.17$\pm$0.25 & 92.73 $\pm$0.21 \\
    MS2AP \cite{Bi2021MS} & $C$, 2021 & 87.91$\pm$0.19 & 90.98$\pm$0.21  \\
    Contourlet CNN \cite{Liu2021CCNN} & $C$, 2021 & 85.93$\pm$0.51 & 89.57$\pm$0.45 \\
    LSENet \cite{Bi2021LSENet} & $C$, 2021 & 91.93$\pm$0.19 & 93.14$\pm$0.15 \\
    MBLANet \cite{Chen2021MBLANet} & $C$, 2021 & 92.32$\pm$0.15 & 94.66 $\pm$0.11 \\
    LiGNet \cite{Xu2021LiGNet} & $C$, 2021 & 90.23$\pm$0.13 & 93.25$\pm$0.12 \\
    DMSMIL \cite{ICASSP2021} & $C$, 2021 & 91.93$\pm$0.16 & 93.05$\pm$0.14\\
    \hline
    \textbf{AGOS} (ResNet-50) & $C$ & 92.47$\pm$0.19 & 94.28$\pm$0.16 \\
    \textbf{AGOS} (ResNet-101) & $C$ & \textcolor{blue}{92.91$\pm$0.17} & \textcolor{blue}{94.69$\pm$0.18}\\
    \textbf{AGOS} (DenseNet-121) & $C$ & \textbf{93.04$\pm$0.35} & \textbf{94.91$\pm$0.17}\\
    \hline
    \end{tabular} 
    \begin{tablenotes}
        \footnotesize
        \item ‘----’: not reported, ‘*’: not reported \& conducted by us
    \end{tablenotes}
    \label{tab3}
    \end{threeparttable}
\end{table}

\begin{enumerate}[(1)]
\item Our AGOS outperforms all the compared state-of-the-art performance when the training ratios are both 10\% and 20\%. Its DenseNet-121 and ResNet-101 version achieves the best and second best results on both settings, while the performance of ResNet-50 version is competitive.
\item Generally speaking, those approaches highlighting the key local regions of an aerial scene \cite{Gong2018When,WangQi2018RA,Bi2019APDC,Bi2020RADC,Bi2019Multiple} or building a multi-scale convolutional feature representation tend to achieve a better performance \cite{He2018Remote,Bi2021MS,Han2017Pre}.
\item The performance of GAN based approaches \cite{Lin2017MARTA,Yu2019AGAN} degrades significantly when compared with other CNN based methods on NWPU. Specifically, they are weaker than some CNN baselines such as VGGNet and GoogLeNet. 
\end{enumerate}

Moreover, the per-category classification accuracy under the training ratio of 10\% and 20\% is shown in Fig.~\ref{fig7}~(e), (f). Most categories of the NWPU dataset are classified well. Similar to the discussion on AID, potential explanations of these outcomes include: 

\begin{enumerate}[(1)]
\item The difference of spatial resolution and object size is more varied in NWPU than in AID and UCM. Thus, the importance of both highlighting the key local regions and building more discriminative multi-grain representation is critical for an approach to distinguish the aerial scenes of different categories. The weak performance of GAN based methods can also be accounted that no effort has been investigated on either of the above two strategies, which is an interesting direction to explore in the future.
\item As our AGOS builds multi-grain representations and highlights the key local regions, it is capable of distinguishing some scene categories that are varied a lot in terms of object sizes and spatial density. Thus, the experiments on all three benchmarks reflect that our AGOS is qualified to distinguish such scene categories. 
\end{enumerate}

\begin{figure*}[!t]
    \centering %插入的图片居中表示
    \includegraphics[width=7.0in]{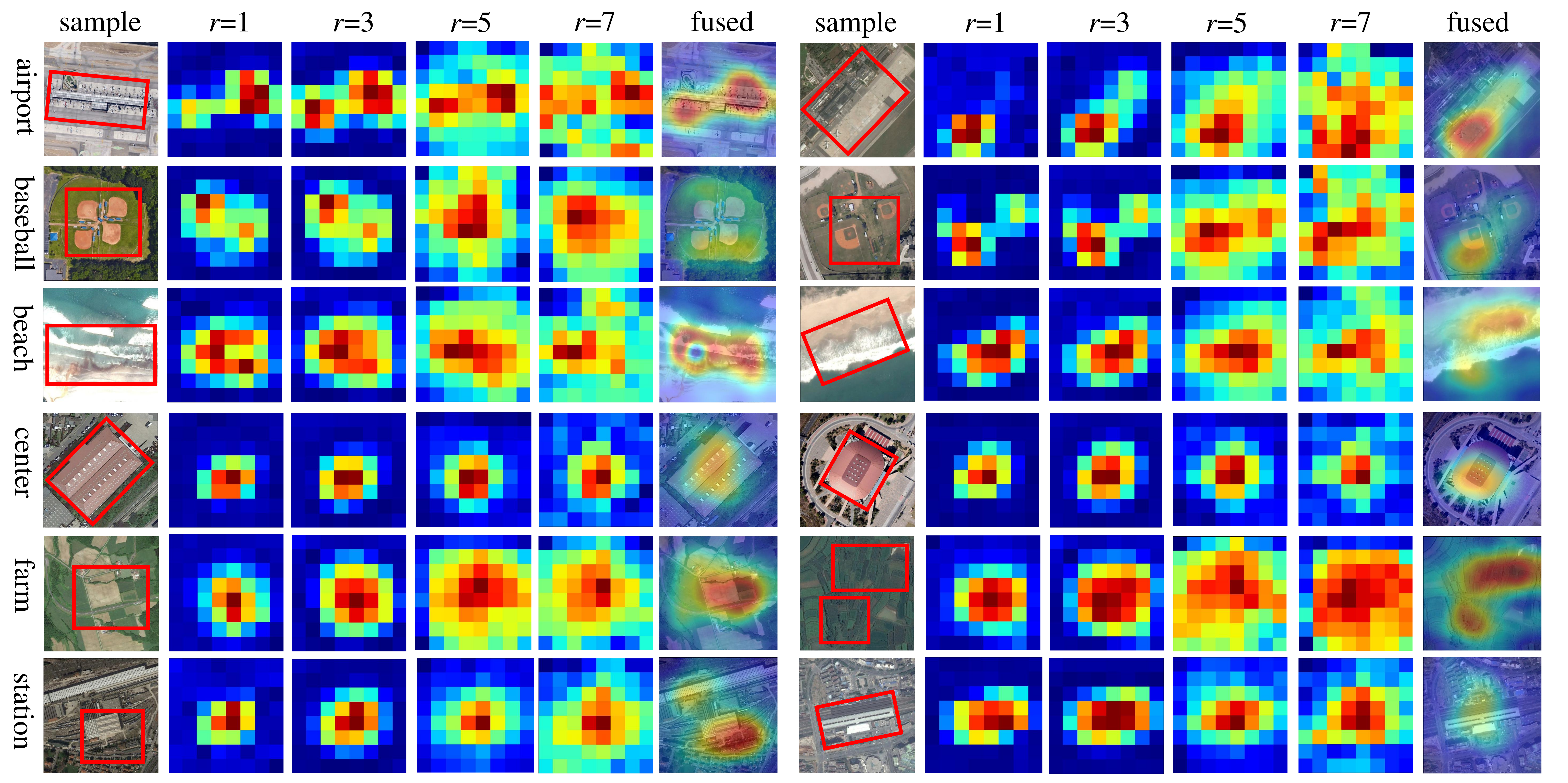}
	\caption{Visualized feature responses by our AGOS (with ResNet-50 backbone) (denoted as \textit{fused}) and the intermediate instance presentation in our DDC when the dilated rate is 1 (denoted as $r=1$), 3 (denoted as $r=3$), 5 (denoted as $r=5$) and 7 (denoted as $r=7$) respectively. For each intermediate instance presentation, the feature responses of each instance on each scene category are averaged and normalized to $[0, 255]$ without resizing. Key local regions of the scene are labelled in red bounding boxes for reference. Heatmaps (fused) are generated from the average of instance representations from all grains and resized to the original image size, and then normalized to $[0, 255]$.}  %图片的名称
	\label{vis}   %标签，用作引用
\end{figure*}

\subsection{Ablation studies}

Apart from the ResNet-50 baseline, our AGOS framework consists of a multi-grain perception (MGP) module, a multi-branch multi-instance representation (MBMIR) module and a self-aligned semantic fusion (SSF) module.
To evaluate the influence of each component on the classification performance, we conduct an ablation study on AID benchmark and the results are reported in Table~\ref{tab4}. It can be seen that:

\begin{table}[!t]  
    \centering
    \caption{Ablation study of our AGOS (with ResNet-50 backbone) on AID dataset; Metrics presented in \%; ResNet: backbone ResNet-50; MGP: multi-grain perception module; MBMIR: multi-branch multi-instance representation module; SSF: self-aligned semantic fusion module; $L_{cls}$: only use the classification term in the loss function.}
    \small
    \begin{tabular}{c|c|c|c|c|c} 
    %\toprule[2pt]
    \hline
    \multicolumn{4}{c|}{Module} & \multicolumn{2}{c}{AID}\\
    \cline{1-6}
    ResNet & MGP & MBMIR & SSF & 20\% & 50\%\\
    \hline
    $\checkmark$ & ~ & ~ & ~ & 88.63$\pm$0.26 & 91.72$\pm$0.17 \\
    \hline
    $\checkmark$ & $\checkmark$ & ~ & ~ & 89.99$\pm$0.21 & 92.98$\pm$0.19 \\
    $\checkmark$ & $\checkmark$ & $\checkmark$ & ~ & 94.16$\pm$0.24 & 96.20$\pm$0.16 \\
    $\checkmark$ & $\checkmark$ & ~ & $\checkmark$ & 92.26$\pm$0.25 & 95.13$\pm$0.15 \\
    $\checkmark$ & $\checkmark$ & $\checkmark$ & $L_{cls}$ & 94.27$\pm$0.19 & 96.47$\pm$0.23 \\
    \hline
    $\checkmark$ & $\checkmark$ &$\checkmark$ & $\checkmark$ & \textbf{94.99$\pm$0.24} & \textbf{97.01$\pm$0.18} \\
    \hline
    %\bottomrule[2pt]
    \end{tabular} 
    \label{tab4}
\end{table}

\begin{enumerate}[(1)]
\item The performance gain led by MGP is about 1.26\% and 1.36\% if directly fused and then fed into the classification layer. Thus, more powerful representation learning strategies are needed for aerial scenes. 
\item Our MBMIR module leads a performance gain of 4.17\% and 3.22\% respectively. Its effectiveness can be explained from: 1) highlighting the key local regions in aerial scenes by using classic MIL formulation; 2) building more discriminative multi-grain representation by extending MIL to the multi-grain form.
\item Our SSF module improves the performance by about 1\% in both two cases. This indicates that our bag scheme self-alignment strategy is effective to further refine the multi-grain representation so that the representation from each grain focuses on the same bag scheme. 
\end{enumerate}

To sum up, MGP serves as a basis in our AGOS to perceive the multi-grain feature representation, and MBMIR is the key component in our MBMIR which allows the entire feature representation learning under the MIL formulation, and the performance gain is the most. Finally, our SSF helps further refine the instance representations from different grains and allows the aerial scene representation more discriminative. 

\subsection{Generalization ability}
\textit{1) On different backbones:} Table~\ref{tab5} lists the classification performance, parameter number and inference time of our AGOS framework when embedded into three commonly-used backbones, that is, VGGNet \cite{VGGNet}, ResNet \cite{ResNet} and Inception \cite{GoogLeNet} respectively. It can be seen that on all three backbones our AGOS framework leads to a significant performance gain while only increasing the parameter number and lowing down the inference time slightly. The marginal increase of parameter number is quite interesting as our AGOS removes the traditional fully connected layers in CNNs, which usually occupy a large number of parameters. 

\begin{table}[!t]  
    \centering
    \caption{Performance of our AGOS on different backbones on AID dataset under the 50\% training ratio \cite{Xia2017AID}; Metric presented in \%; Para. num.: Parameter numbers; presented in million; FPS: Frame Per Second.}
    \small
    \begin{tabular}{c|c|cc} 
    %\toprule[2pt]
    %\hline
    \cline{1-4}
    ~ & OA & Para. num. & FPS \\
    \hline
    VGG-16 &  90.64$\pm$0.14 & 15.43 & 245.82 \\
    %\hline
    AGOS with VGG-16 & \textbf{96.26$\pm$0.15} & 19.94 & 227.48 \\
    \hline
    ResNet-50 & 91.72$\pm$0.17 & 23.46 & 422.30 \\
    AGOS with ResNet-50 & \textbf{97.01$\pm$0.18} & 29.69 & 367.40 \\
    \hline
    Inception-v2 & 91.40$\pm$0.19 & 6.64 & 704.22 \\
    AGOS with Inception-v2 & \textbf{96.64$\pm$0.16} & 8.79 & 652.35 \\
    \hline
    %\bottomrule[2pt]
    \end{tabular} 
    \label{tab5}
\end{table}

\textit{2) On classification task from other domains:} Table~\ref{tab6} reports the performance of our AGOS framework on a medical image classification \cite{LAG} and a texture classification \cite{KTD} benchmark respectively. The dramatic performance gain compared with the baseline on both benchmarks indicates that our AGOS has great generalization capability on other image recognition domains.

\begin{table}[!t]  
    \centering
    \caption{Performance of our AGOS framework (with ResNet-50 backbone) on LAG \cite{LAG} and KTD dataset \cite{KTD}; Both benchmarks require the five-fold classification accuracy; Metric presented in \%.}
    \small
    \begin{tabular}{c|c|c} 
    %\toprule[2pt]
    %\hline
    \cline{1-3}
    ~ & LAG & KTD \\
    \hline
    ResNet & 91.75 & 91.74 \\
    %\hline
    AGOS with ResNet & \textbf{98.05} & \textbf{99.95}\\
    %~ & \textcolor{red}{$\uparrow$ 6.87\%} & \textcolor{red}{$\uparrow$ 8.95\%}\\
    \hline
    %\bottomrule[2pt]
    \end{tabular} 
    \label{tab6}
    \vspace{-1.0em}
\end{table}

\subsection{Discussion on bag scheme alignment}
Generally speaking, the motivation of our self-aligned semantic fusion (SSF) module is to learn a discriminative aerial scene representation from multi-grain instance-level representations. However, in classic machine learning and statistical data processing, there are also some solutions that either select or fit an optimal outcome from multiple representations. Hence, it would be quite interesting to compare the impact of our SSF and these classic solutions. 

To this end, four classic implementations on our bag probability distributions from multi-grain instance representations, namely, naive mean (Mean) operation, naive max (Max) selection, majority vote (MV) and least squares method (LS), are tested and compared based on the AID dataset under the 50\% training ratio. Table.~\ref{tabcompu} lists all these results. Note that by using naive mean the entire framework degrades to the third case in our ablation studies (in Table.~\ref{tab4}). 

\begin{table}[!t]  
    \centering
    \caption{Comparison of our AGOS (with ResNet-50 backbone) with some classic solutions on aligning the aerial scene scheme on AID benchmark; Mean: mean operation; Max: max selection; MV: majority vote; LS: least squares method; Metric presented in \%.}
    \small
    \begin{tabular}{c|c} 
    %\toprule[2pt]
    %\hline
    \cline{1-2}
    Method & AID 50\% \\
    \hline
    Mean & 96.20$\pm$0.16 \\
    Max & 95.94$\pm$0.21 \\
    MV & 96.43$\pm$0.17 \\
    LS & 96.38$\pm$0.19 \\
    \hline
    \textbf{AGOS} (ours) & \textbf{97.01$\pm$0.18}\\
    \hline
    %\bottomrule[2pt]
    \end{tabular} 
    \label{tabcompu}
    \vspace{-1.0em}
\end{table}

It can be seen that our SSF achieves the best performance while: 1) max selection shows apparent performance decline; 2) other three solutions, namely mean operation, majority vote and least square, do not show much performance difference.

\begin{figure*}[!t]
    \centering %插入的图片居中表示
    \subfigure[mean selection]{
    \includegraphics[width=1.25in]{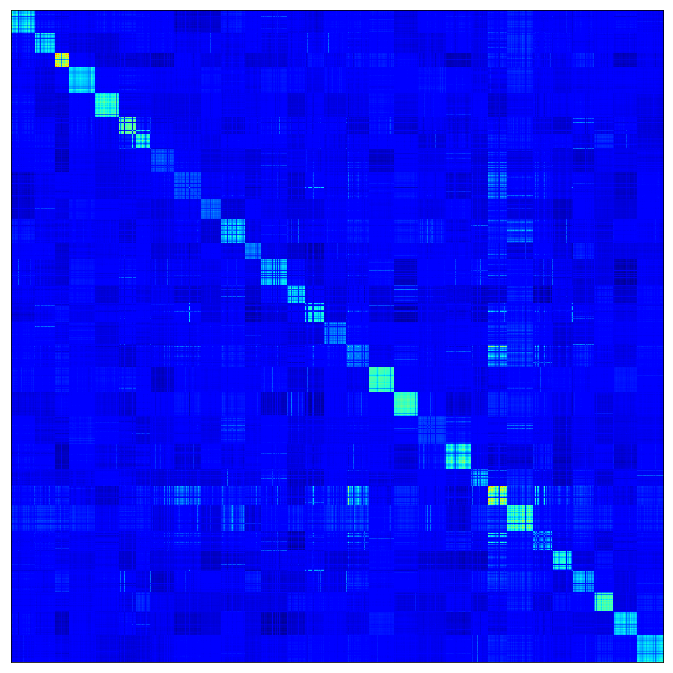}
    }
    \subfigure[max selection]{
    \includegraphics[width=1.25in]{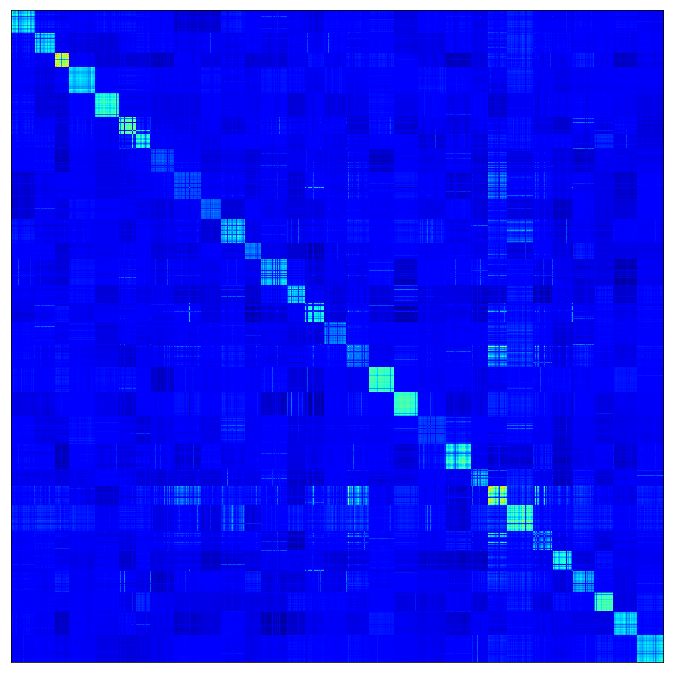}
    }
    \subfigure[majority vote]{
    \includegraphics[width=1.25in]{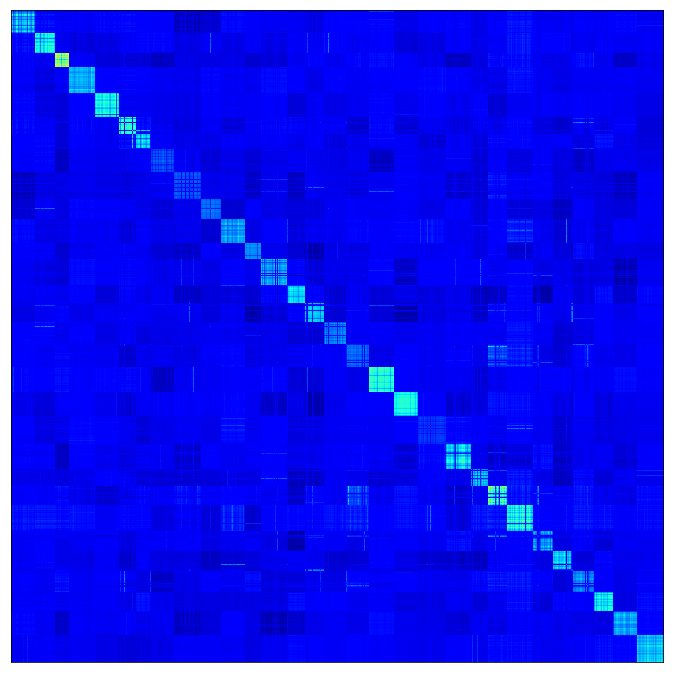}
    }
    \subfigure[least square]{
    \includegraphics[width=1.25in]{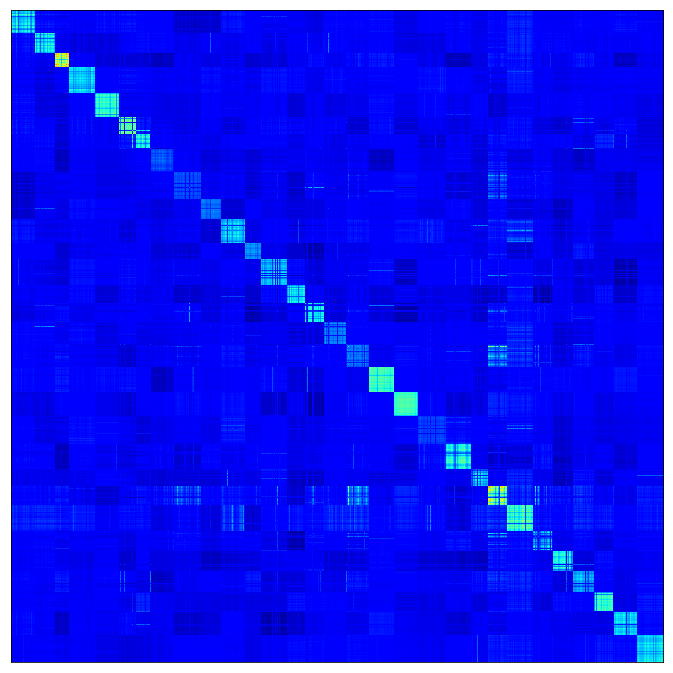}
    }
    \subfigure[AGOS (ours)]{
    \includegraphics[width=1.25in]{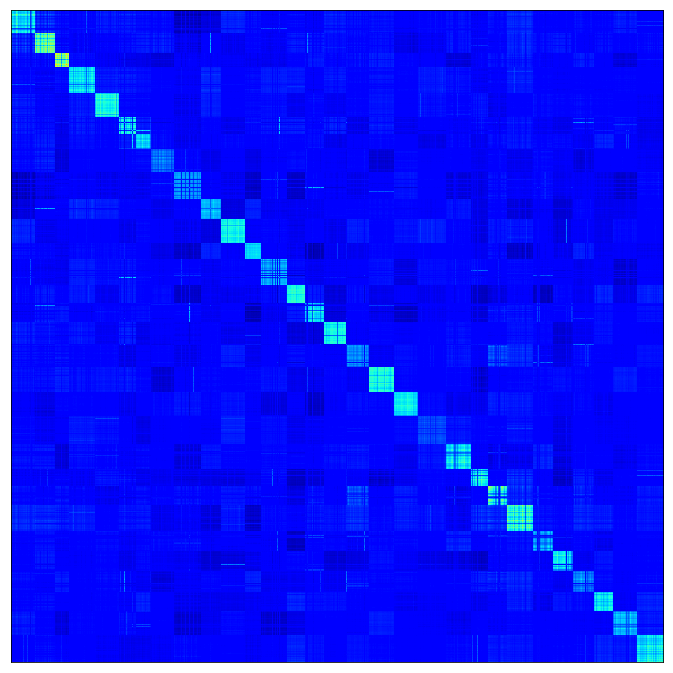}
    }
	\caption{Visualized co-variance matrix of the bag probability distribution after scene scheme alignment, processed by mean selection (a), max selection (b), majority vote (c), least square method (d) and our AGOS (e). Ideally, the co-variance matrix of bag probability distribution should have high responses in the diagonal region and no responses in other regions. }  
	\label{covvis}   
\end{figure*}

To better understand how these methods influence the scene scheme alignment, Fig.~\ref{covvis} offers the visualized co-variance matrix of the bag probability distributions from all the test samples. Generally speaking, a good scene representation will have higher response on the diagonal region while the response from other regions should be as low as possible. It is clearly seen that our SSF has the best discrimination capability, while for the other solutions some confusion between bag probability distributions of different categories always happens.

The explanation may lie in the below aspects: 1) Our SSF aligns the scene scheme from both representation learning and loss optimization, and thus leads to more performance gain; 2) naive average on these multi-grain instance representations already achieves an acceptable scene scheme representation, and thus leaves very little space for other solutions such as least square and majority vote to improve; 3) max selection itself may lead to more variance on bag probability prediction and thus the performance declines.

\subsection{Sensitivity analysis}
\subsubsection{Influence of grain number} 
\begin{figure}[!t]
    \centering %插入的图片居中表示
    \subfigure[AID 20\%]{
    \includegraphics[width=1.6in]{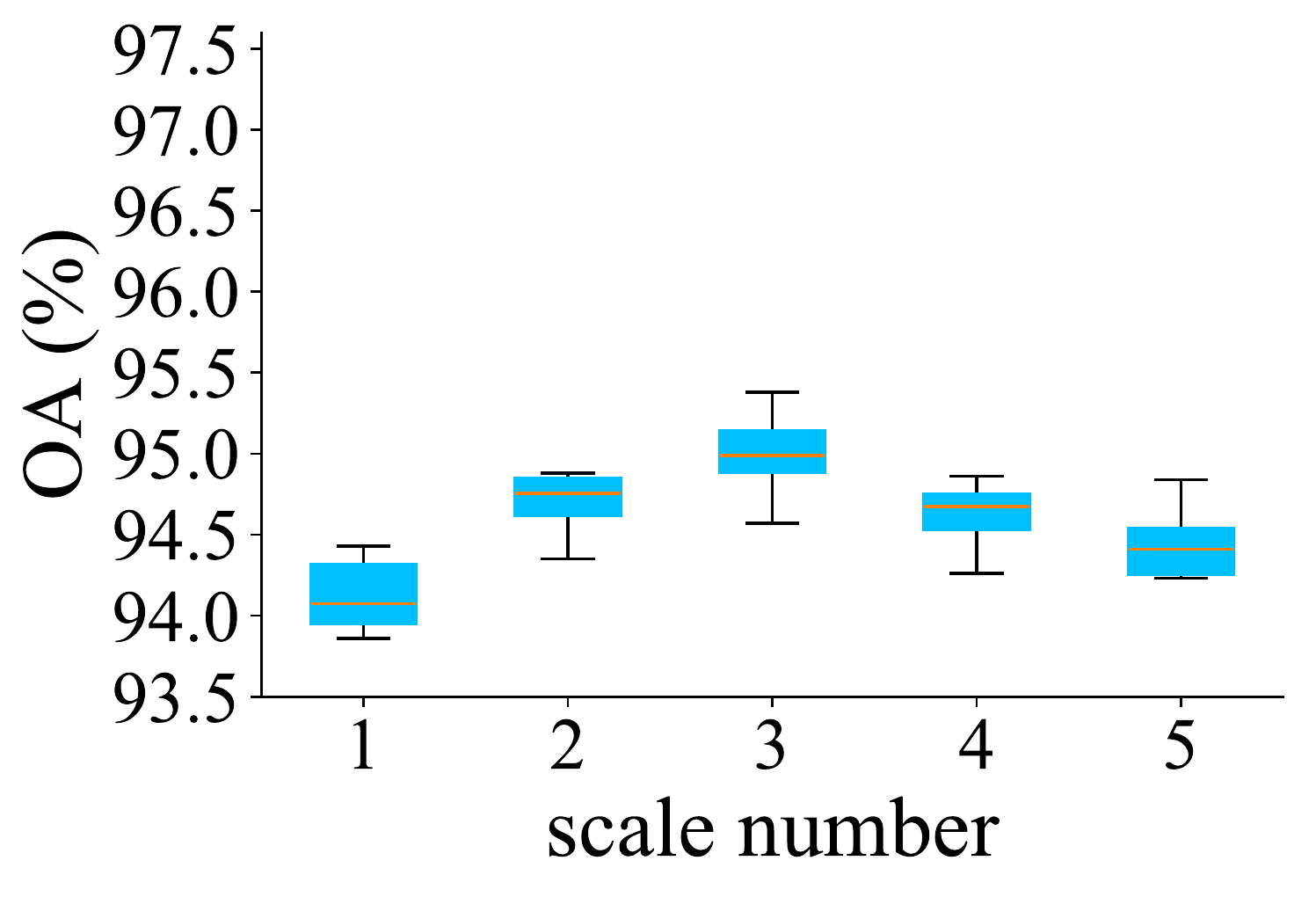}
    }
    \subfigure[AID 50\%]{
    \includegraphics[width=1.6in]{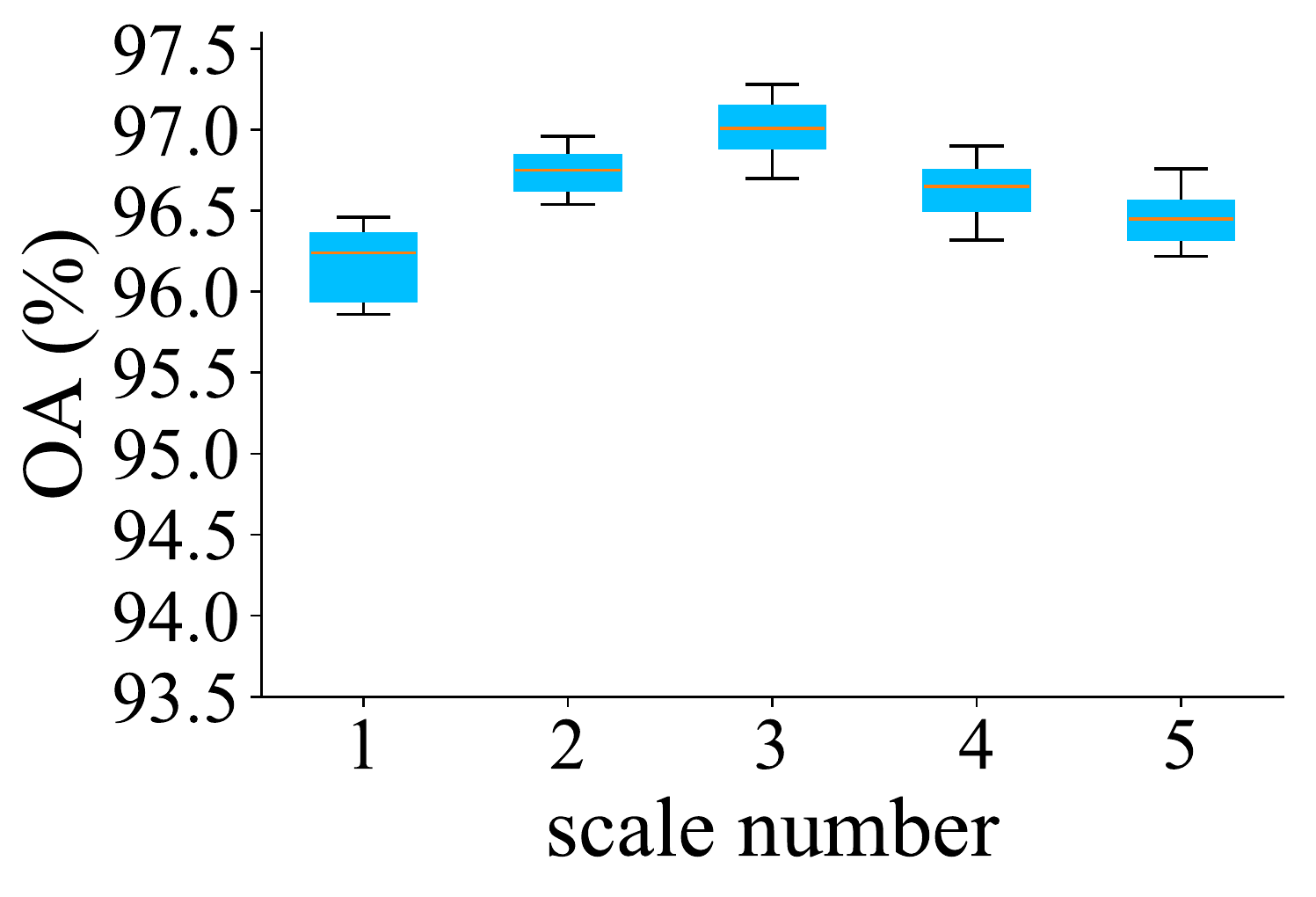}
    }
     %插入的图，包括JPG,PNG,PDF,EPS等，放在源文件目录下
	\caption{Performance change influenced by the grain number in our AGOS (with ResNet-50 backbone) on AID 20\% (a) and 50\% (b).}  %图片的名称
	\label{fig8}   %标签，用作引用
\end{figure}

Fig~\ref{fig8}~lists the performance when the grain number $T$ in our AGOS changes. It can be seen that when there are about 3 or 4 grains, the classification accuracy reaches its peak. After that, the classification performance slightly declines. This implies that the combined utilization of convolutional features when the dilated rate is 1, 3 and 5 is most discriminative in our AGOS. When there are too many grains, the perception field becomes too large and the scene representation becomes less discriminative. Also, when the grain number is little, the representation is not qualified enough to completely depict the semantic representation where the key objects vary greatly in sizes. 

On the other hand, the visualized samples in Fig.~\ref{vis}~also reveal that when the dilation rate in our MGP is too small, the instance representation tends to focus on a small local region of an aerial scene. In contrast, when the dilation rate is too large, the instance representation activates too many local regions irrelevant to the scene scheme. Thus, the importance of our scene scheme self-align strategy reflects as it helps the representation from different grains to align to the same scene scheme and refines the activated key local regions. Note that for further investigating the interpretation capability of these patches and the possibility for weakly-supervised localization task, details can be found in ~\cite{Shi2021loss}. 

\subsubsection{Influence of hyper-parameter $\alpha$} 
\begin{figure}[!t]
    \centering %插入的图片居中表示
	 \subfigure[AID 20\%]{
    \includegraphics[width=1.6in]{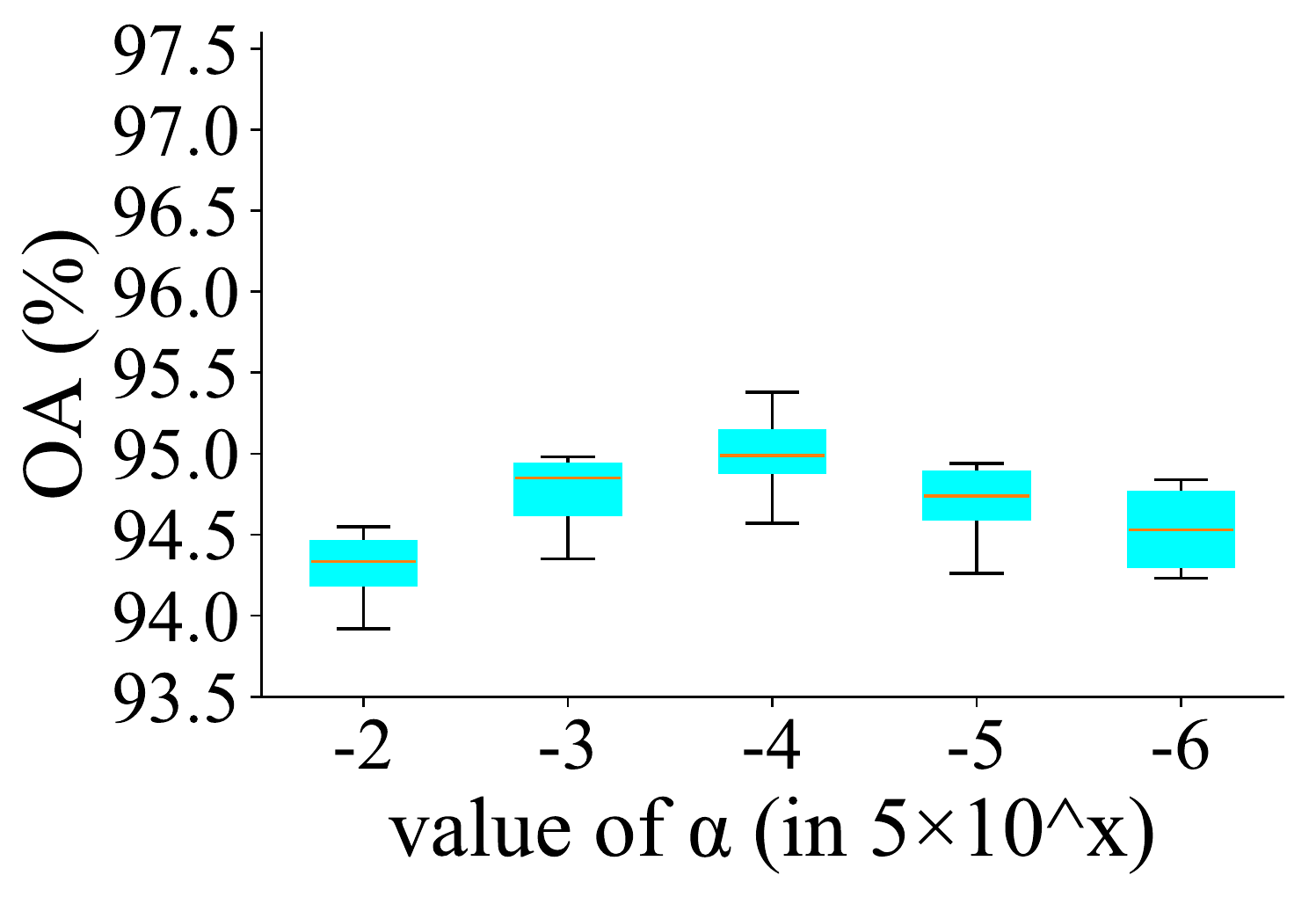}
    }
    \subfigure[AID 50\%]{
    \includegraphics[width=1.6in]{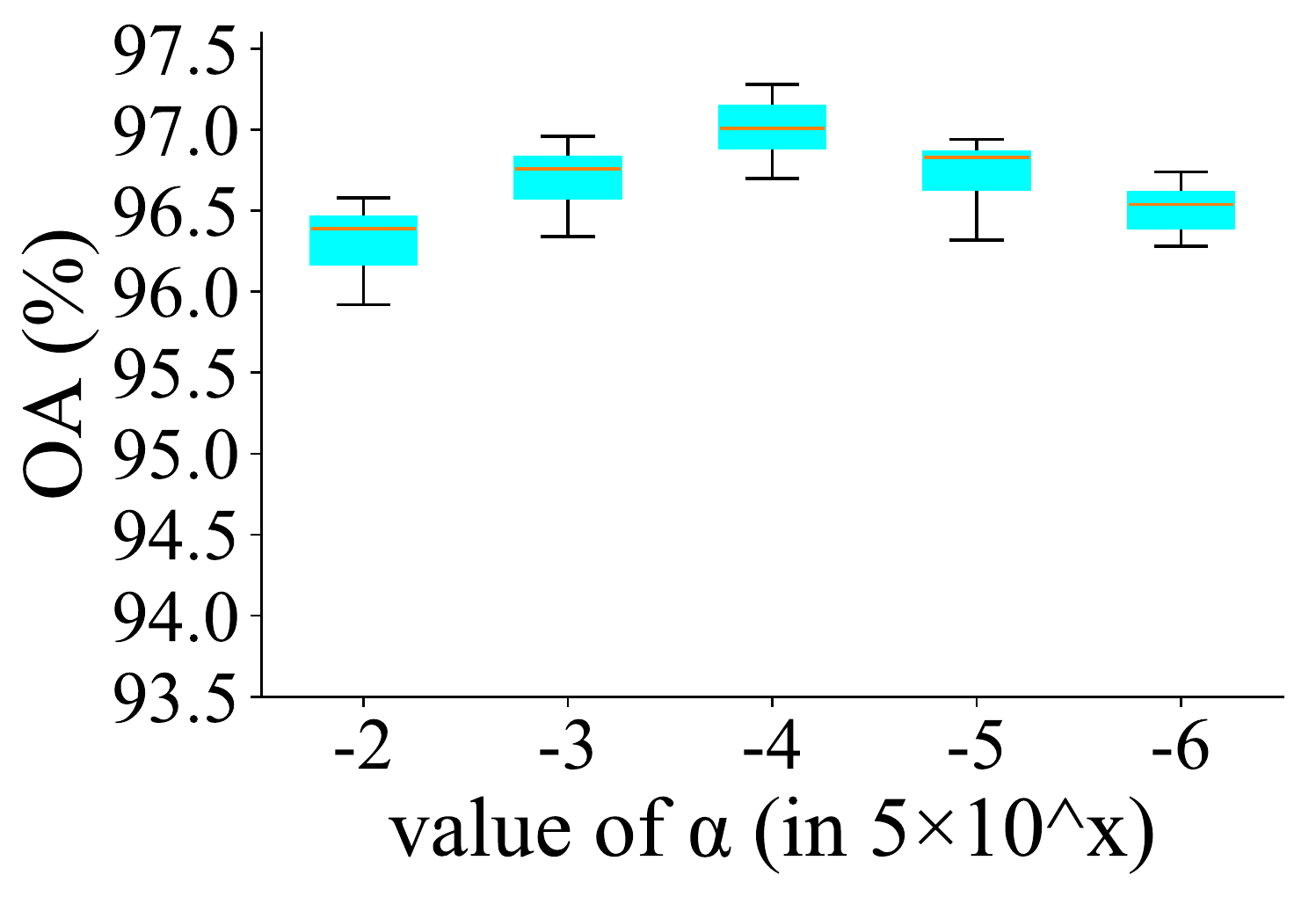}
    } %插入的图，包括JPG,PNG,PDF,EPS等，放在源文件目录下
	\caption{Performance change influenced by the hyper-parameter $\alpha$ in our AGOS (with ResNet-50 backbone) on AID 20\% (a) and 50\% (b); $\alpha$ presented in $5\times10^{x}$, and $x=-2,-3,-4$ and $-5$.}  %图片的名称
	\label{fig9}   %标签，用作引用
\end{figure}

Fig.~\ref{fig9}~shows the classification accuracy fluctuation when the hyper-parameter $\alpha$ in our loss function changes. It can be seen that the performance of our AGOS is stable when $\alpha$ changes. However, when it is too large, the performance shows an obvious decline. When it is too small, the performance degradation is slight. 

\subsubsection{Influence of differential dilated convolution}
Table~\ref{tab8}~lists the classification performance when every component of differential dilated convolution (DDC) in our MGP is used or not used. It can be seen that both the differential operation (D\#DC) and the dilated convolution (DD\#C) lead to an obvious performance gain for our AGOS. Generally, the performance gain led by the dilated convolution is higher than the differential operation as it enlarges the receptive field of a deep learning model and thus enhances the feature representation more significantly. 

\begin{table}[!t]  
    \centering
    \caption{Comparison of our differential dilated convolution (DDC) on the cases when not using differential operation (D\#DC), not using dilated convolution (DD\#C) and not using either differential operation and dilated convolution (C) on AID benchmark with ResNet-50 backbone; Metric in \%.}
    \small
    \begin{tabular}{c|cc} 
    %\toprule[2pt]
    %\hline
    \cline{1-3}
    ~ & AID 20\% & AID 50\% \\
    \hline
    C & 90.56$\pm$0.25 & 93.45$\pm$0.19 \\
    DD\#C & 91.34$\pm$0.21 & 94.36$\pm$0.16 \\
    D\#DC & 93.85$\pm$0.22 & 96.18$\pm$0.17 \\
    \hline
    DDC & \textbf{94.99$\pm$0.24} & \textbf{97.01$\pm$0.18}\\
    \hline
    %\bottomrule[2pt]
    \end{tabular} 
    \label{tab8}
    \vspace{-1.0em}
\end{table}

\section{Conclusion}
\label{sec5}
In this paper, we propose an \textit{all grains, one scheme} (AGOS) framework for aerial scene classification. \textit{To the best of our knowledge}, it is the first effort to extend the classic MIL into deep multi-grain MIL formulation. The effectiveness of our AGOS lies in three-folds: 1) The MIL formulation allows the framework to highlight the key local regions in determining the scene category; 2) The multi-grain multi-instance representation is more capable of depicting the complicated aerial scenes; 3) The bag scheme self-alignment strategy allows the instance representation from each grain to focus on the same bag category. 
Experiments on three aerial scene classification datasets demonstrate the effectiveness of our AGOS and its generalization capability. 

As our AGOS is capable of building discriminative scene representation and highlighting the key local regions precisely, our future work includes transferring our AGOS framework to other tasks such as object localization, detection and segmentation especially under the weakly-supervised scenarios.

\section*{Acknowledgment}
This research is partially supported by National Natural Science Foundation of China under contracts No.U2033216 and No.U1833201, and foundation from Key Laboratory of National Geographic Census and Monitoring, Ministry of Nature Resources under contract No.2020NGCMZD03. This research is also funded by the National Natural Science Foundation of China under contracts No.61922065 and No.61771350, No.41820104006 and partially funded by the Science and Technology Major Project of Hubei Province (Next-Generation AI Technologies) under Grant 2019AEA170 and by the Shanghai Aerospace Science and Technology Innovation Project under Grant SAST2019-094.

%The authors would like to thank the editors and the anonymous four reviewers, whose insightful suggestions and comments significantly improved our manuscript.

\bibliographystyle{ieeetr} %ieeetr国际电气电子工程师协会期刊
\bibliography{cameraready} % ref

\end{document}